\documentclass[preprint,12pt]{elsarticle}
\biboptions{sort&compress}
\usepackage{float} 

\usepackage{amssymb}
\usepackage{amsmath}
\usepackage{booktabs}
\usepackage{multirow}
\usepackage{lineno}
\usepackage{tabularx}
\newcolumntype{b}{X}
\newcolumntype{s}{>{\hsize=.5\hsize}X}
\usepackage{subcaption}
\journal{Nuclear Physics B}

\begin{document}

\begin{frontmatter}



\title{TrajGATFormer: A Graph-Based Transformer Approach for Worker and
Obstacle Trajectory Prediction in Off-site Construction Environments}

 \author[label1]{Mohammed Alduais}
 
 \affiliation[label1]{organization={Department of Civil \& Environmental Engineering, University of Alberta},
            addressline={}, 
            city={Edmonton},
            postcode={T6G 2R3}, 
            state={AB},
            country={Canada}}
 \author[label2]{Xinming Li}
 \affiliation[label2]{organization={Department of Mechanical Engineering, University of Alberta},
            addressline={}, 
            city={Edmonton},
            postcode={T6G 2R3}, 
            state={AB},
            country={Canada}}

\author[label1]{Qipei Mei\corref{cor1}}
\cortext[cor1]{Corresponding author's email address: qipei.mei@ualberta.ca}


\begin{abstract}
As the demand grows within the construction industry for processes that are not only faster but also safer and more efficient, offsite construction has emerged as a solution, though it brings new safety risks due to the close interaction between workers, machinery, and moving obstacles. Predicting the future trajectories of workers and taking into account social and environmental factors is a crucial step for developing collision-avoidance systems to mitigate such risks. Traditional methods often struggle to adapt to the dynamic and unpredictable nature of construction environments. Many rely on simplified assumptions or require hand-crafted features, limiting their ability to respond to complex, real-time interactions between workers and moving obstacles. While recent data-driven methods have improved the modeling of temporal patterns, they still face challenges in capturing long-term behavior and accounting for the spatial and social context crucial to collision risk assessment. To address these limitations, this paper proposes a comprehensive framework integrating YOLOv10n and DeepSORT for precise detection and tracking, alongside two novel trajectory prediction models: TrajGATFormer and TrajGATFormer-Obstacle. YOLOv10n serves as the backbone for object detection, reliably identifying workers and obstacles in diverse construction scenes, while DeepSORT efficiently tracks each detected entity over time, assigning unique IDs to maintain continuity across frames. The two predictive models leverage a transformer encoder-decoder architecture with Graph Attention Networks (GAT) to capture both temporal dependencies and spatial interactions. TrajGATFormer focuses on predicting worker trajectories, achieving an average displacement error (ADE) of 1.25 meters and a final displacement error (FDE) of 2.3 meters over a 4.8-second prediction horizon. TrajGATFormer-Obstacle extends this framework to handle both worker and obstacle predictions, achieving even higher accuracy, with an ADE of 1.15 meters and an FDE of 2.2 meters. Comparative analysis showed that both models outperformed traditional methods, reducing ADE and FDE by up to 35\% and 38\%, respectively. These models can serve as the backbone for constructing collision-avoidance alarm systems that can be installed at offsite construction facilities.
\end{abstract}




\end{frontmatter}



\section{Introduction}
  The construction industry is considered as one of the major drivers of the global economy in the coming few years. According to a report by Oxford Economics, the global construction industry is projected to grow by 42\% this decade, reaching 15.2 trillion USD compared to 10.7 trillion USD in the previous decade \cite{oxford2021future}. Furthermore, total global spending on construction accounted for 13\% of global GDP in 2020 and is projected to rise to 13.5\% by 2030 \cite{oxford2021future}. Although the construction sector contributes significantly to economic growth, it is also one of the most hazardous industries. In 2020 alone, the construction industry in the United States was responsible for 1,034 fatal injuries and more than 70,000 non-fatal injuries \cite{harris2023focus}. Between 2011 and 2021, more than 10,000 construction workers were killed on the job due to exposure to various hazardous situations, including (1) falls to lower levels, (2) being struck by objects, (3) electrocution, (4) caught-in/between incidents, and (5) other hazards \cite{harris2023focus}. According to OSHA, falls to lower levels, being struck by objects, electrocution, and caught-in/between incidents are classified as "Focus Four Hazards," whereas all other hazards are classified as "Non-Focus Four Hazards" \cite{focusfour2019}. Focus Four hazards are the primary cause of fatal injuries, accounting for up to 65.5\% of all fatal incidents, which corresponds to approximately 6,900 deaths. Similarly, from 2011 to 2020, Focus Four injuries accounted for over 40.4\% of non-fatal injuries, totaling more than 315,000 incidents. Struck-by accidents accounted for more than 17\% of fatal injuries, accounting for 1,700 injuries, and more than 20\% non-fatal injuries, accounting for more than 167,000 injuries.
  \par
    As a result, off-site construction emerges as a possible approach to improve worker safety. Off-site construction provides a faster, safer and more controlled working environment with improved quality control practices, and is less harmful to the environment compared to traditional on-site construction \cite{lawson1999modular,yu2018rigidity, stern2017steel, jin2019environmental}. According to the Mackenzie report, off-site construction can reduce construction time by 50\% and reduce cost in some projects by 20\% \cite{mckinsey2019modular}. However, off-site construction facilities are dynamic environments in which workers operate in congested areas and in close proximity to construction equipment such as lifting cranes and robotic machines, which can lead to unsafe situations \cite{teizer2015proximity}. Furthermore, the lack of positional awareness among workers when operating construction machinery can be attributed to abrupt movements and rotations of the machinery, obscured sight lines, and the noisy working environment \cite{teizer2015proximity}. Therefore, to mitigate such risks, the prediction of the worker's trajectory, which can be defined as the prediction of future positions based on their previous observations and social and environmental interactions \cite{Korbmacher2022}, plays a crucial role in providing essential information for the development of a struck-by alarm system that alerts workers when a potential hazard situation can occur \cite{buildings13061502,Cai2020}.
    
    Recent advances in trajectory prediction have introduced various approaches that range from traditional statistical models to sophisticated machine learning techniques \cite{Korbmacher2022}. Traditional methods, like Kalman Filters (KF) \cite{10.1115/1.3662552}, and Hidden Markov Models (HMMs) \cite{18626}, rely on predefined assumptions and historical data, which often limit their adaptability in highly dynamic and unpredictable settings like construction sites \cite{Korbmacher2022}. More recently, data-driven approaches, such as Long Short-Term Memory (LSTM) networks \cite{10.1162/neco.1997.9.8.1735}, have become popular for their ability to capture temporal patterns in sequential data. LSTMs have shown promise in predicting worker movement by leveraging historical and contextual information, they still struggle in complex environments that require accurate long-term predictions \cite{app11178129}.

    Transformer models have gained attention for trajectory prediction due to their inherent capability to capture complex dependencies across long time-frames, without the limitations of traditional recurrent models\cite{vaswani2017attention,FRANCO2023109372}. While their initial success was demonstrated in natural language processing, their versatility has made them popular in various domains, including trajectory prediction \cite{giuliari2020transformernetworkstrajectoryforecasting}. Additionally, Graph Attention Networks (GATs) have been applied to model spatial relationships between entities \cite{9010834}, making them well-suited for dynamic environments where interactions between workers and machinery are crucial. Combining transformers with GATs offers a comprehensive approach to trajectory prediction by accounting for both temporal and spatial interactions.

    In this paper, we propose a trajectory prediction framework that includes two distinct models. The first model is designed specifically for predicting worker trajectories, while the second model extends this capability by incorporating both workers and moving obstacles. The framework begins by detecting and tracking objects using a surveillance camera, followed by preprocessing the acquired data. Each model employs a hybrid approach that combines Transformer networks and GATs. By leveraging the strengths of both architectures, the models effectively capture the spatial and temporal complexities of dynamic environments, ensuring more accurate trajectory predictions.

    The paper is organized as follows: Section \ref{Related Work}  reviews the literature and methods used for trajectory prediction. Section \ref{section:Proposed framework} introduces the proposed framework, proposed model architectures, and the implementation details. Section \ref{section:Datasets} discusses the datasets used. Section \ref{Results} presents the results of the proposed models along with different trajectory prediction methods. Section \ref{Conclusions} provides a conclusion, and Section \ref{limitations} presents limitations of the proposed framework and suggests future direction.

\section{Related Work}\label{Related Work}
    With the rapid development in the fields of computer vision and autonomous vehicles, trajectory prediction emerges as one of the fundamental challenges that need to be solved and studied \cite{Cai2020}. There are two approaches to solving the problem of trajectory prediction. 
    \par 
    The first approach is the knowledge-based approach \cite{Korbmacher2022}, which is referred to by multiple names in the literature, for example physics-based \cite{Tordeux2020}, expert-based \cite{Cheng2020} , reasoning-based \cite{Kruse2013}, or traditional approach \cite{Bighashdel2019}. In this approach, models need predefined rules or functions by humans that try to interpret or represent the physical, social, and physiological information of the target \cite{Korbmacher2022}. For example, Constant-Velocity method, is considered as one of the simplest yet effective models for trajectory prediction \cite{schöller2020constantvelocitymodelteach}. This method assumes that motion of vehicles/pedestrians continues without taking into consideration any other factors such as social interactions. Kalman Filtering (KF) \cite{10.1115/1.3662552} which was introduced in 1960, can be used to introduce uncertainty information to kinematic models \cite{Karle_2023}. For example, Elnagar \cite{1013236}, proposed an algorithm to predict the trajectories of moving obstacles combining the KF and the constant acceleration model. However, one of the main disadvantages of these methods is the assumption of constant motion, which results in poor performance when considering long term predictions (\( > 2s\)) \cite{Karle_2023}. Hidden Markov Models (HMMs) are recognized as probabilistic models used for the analysis of sequential data \cite{18626}. These models operate under the assumption that the data follows a Markov process with states that are not directly observable. A model proposed by Ye et al. \cite{ye}, used HMMs with two layers of hidden states for vehicle trajectory prediction. The authors used statistical methods to determine the HMM parameters. In the prediction phase, the authors introduced the Viterbi algorithm to solve the decoding of HMM and find the optimal sequence of hidden states. The social force model (SFM) proposed by Helbing and Molnar \cite{Helbing_1995}, was one of the pioneering works to model pedestrian interaction and inspired many researches to build on this concept. This approach describes that the movement of pedestrians is based on a social force that drives them to perform certain actions, in this case their movement.
    \par
    The second approach is the data-driven approach, also known as supervised deep learning methods \cite{Korbmacher2022}.This approach is widely used in fields such as self-driving cars, robotics, and pedestrian trajectory prediction \cite{Korbmacher2022}. Unlike the knowledge-based approach, this method does not require manual definition of human behavior. Instead, it offers more flexible models capable of adapting to complex scenarios, although it relies on large amounts of data for training \cite{Korbmacher2022}. An example of data-driven approach is Convolution Neural Networks (CNNs), which is a famous method used in computer vision problems such as image classification \cite{ImgeNet}. A model proposed by Nikhil and Morris \cite{nikhil2018convolutionalneuralnetworktrajectory}, where they used a CNN-based model to predict the trajectories of pedestrians. They used a fully connected layer to embed the trajectories, followed by multiple layers of CNN that are used to extract the temporal information. A fully connected layer was added to the end of the model architecture that takes the features from the CNN layers to generate future trajectories. However, this model did not consider social and spatial information for its prediction. To overcome this issue LSTM methods are introduced, which was used by many researchers in the application of trajectory prediction \cite{Korbmacher2022}. One of the pioneers in this field is the Social-LSTM model that was proposed by Alahi et al. \cite{7780479}. The main break throw in his proposed method was the introduction of a social pooling layer that takes into account the influence of the nearby pedestrians. With the success of Social-LSTM, LSTM gained popularity among researchers to use it to solve trajectory prediction. Graph Neural Networks (GNNs) on the other hand have been used in various applications, such as road traffic prediction \cite{li2017diffusion,cui2018high}, text classification \cite{kipf2017semisupervisedclassificationgraphconvolutional}, and in machine translation \cite{vaswani2017attention,shaw2018self,gulcehre2018hyperbolic}. In the domain of trajectory prediction, Social BiGAT which was proposed by Kosaraju et al. \cite{kosaraju2019socialbigatmultimodaltrajectoryforecasting}, Incorporated Graph Attention Networks (GATs) to enhance the ability of the model to capture the social interactions between pedestrians, while the main model architecture uses GAN. However, their limitation lies in the inability to capture temporal dependencies, which are essential for predicting movement over time \cite{9010834}. Since Transformer Networks are capable of capturing long-range dependencies in sequential problems \cite{FRANCO2023109372}, it was utilized as a possible alternative. One of the pioneering works that introduced transformer networks to pedestrian trajectory prediction was the trajectory transformer (TF) by Giuliari et al.  \cite{giuliari2020transformernetworkstrajectoryforecasting}. They used the same transformer encoder-decoder \cite{vaswani2017attention}, which can predict future trajectories of multiple pedestrians at once and reported promising results compared to other trajectory prediction models. However, their approach models each pedestrian without considering any social or scene interactions. 
    \par There are several studies that proposed trajectory predication methods in the construction industry. For example, a model was presented by Zhu et al. \cite{ZHU201695} that was designed to predict workers and moving equipment trajectories based on inputs from two video cameras. This model was designed by utilizing Kalman filters and relies on past trajectories without taking into account other important information, such as the influence of other entities on each other. Wang et al. \cite{wang2019predicting} used a deep region-based convolutional neural network (R-CNN) to detect and classify workers and construction equipment and then designed a CNN-based model for the prediction of the trajectory of these objects. Later, Cai et al. \cite{Cai2020} proposed context-augmented LSTM model that predicts workers trajectories in dynamic construction sites by integrating individual movement with contextual information. However, one of the drawbacks that faced this model that they considered the final destination as a prior knowledge. Another models proposed by Yang et al. \cite{buildings13061502}, proposed an environment-aware worker trajectory prediction model in a modular construction site, using LSTM, and it takes into account the following: (1) worker movement, (2) worker-to-worker and (3) environment-to-worker interactions. However, these models did not consider the influence of dynamic obstacles but considered the use of stationary obstacles as a fixed coordinates. See Table \ref{tab1} for a summary of some of the trajectory prediction methods.
    \par

\begin{table}[ht]
\caption{A Summary of Trajectory Prediction Categories and Their Corresponding Methods\label{tab1}}
\centering
\resizebox{\textwidth}{!}{%
    \begin{tabular}{lccc}
    \toprule
    \textbf{Category} & \textbf{Method} & \textbf{Input} & \textbf{Reference} \\ 
    \toprule
    \multirow{4}{*}{Knowledge-Based} 
        & Constant-Velocity         & Past trajectories                       & \cite{schöller2020constantvelocitymodelteach} \\ \cmidrule(lr){2-4}
        & Kalman filter             & Past trajectories and velocity           & \cite{1013236} \\ \cmidrule(lr){2-4}
        & Hidden Markov Models       & Vehicle past trajectories                & \cite{ye} \\ \cmidrule(lr){2-4}
        & Social Force Model         & Past and neighbor trajectories           & \cite{Helbing_1995} \\ 
    \midrule
    \multirow{4}{*}{Data-driven} 
        & Convolutional Neural Networks  & Displacement volume                    & \cite{nikhil2018convolutionalneuralnetworktrajectory} \\ \cmidrule(lr){2-4}
        & Recurrent Neural Networks       & Past and neighbor trajectories         & \cite{7780479} \\ \cmidrule(lr){2-4}
        & Graph Neural Networks           & Past trajectories and scene information & \cite{kosaraju2019socialbigatmultimodaltrajectoryforecasting} \\ \cmidrule(lr){2-4}
        & Transformer Networks            & Past trajectories                     & \cite{giuliari2020transformernetworkstrajectoryforecasting} \\ 
    \bottomrule
    \end{tabular}%
}
\end{table}

\section{Proposed Framework}\label{section:Proposed framework}
\begin{figure}[!h]
        \centering
        \includegraphics[width=\textwidth]{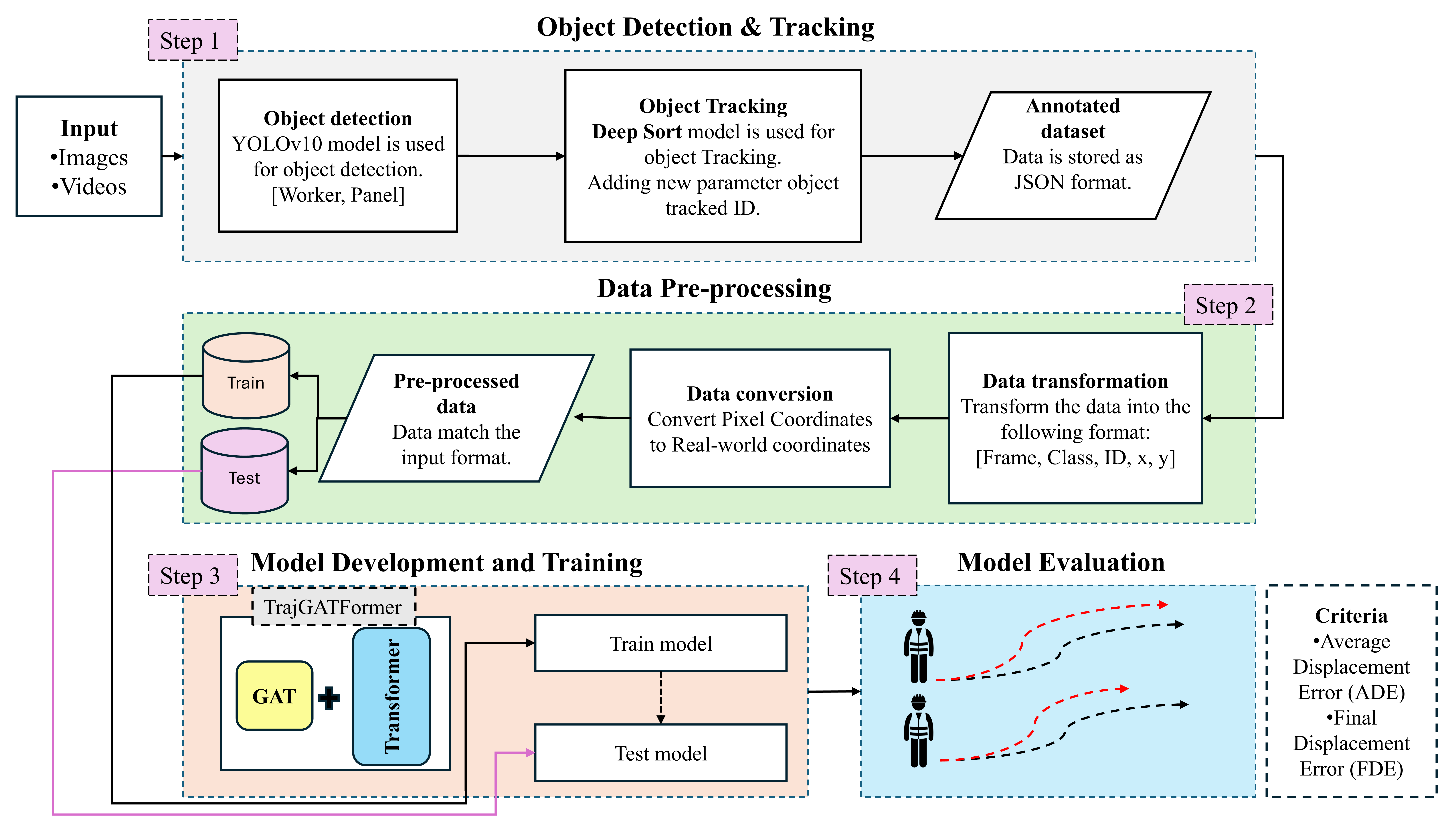}
        \caption{Overview of the proposed framework.}
        \label{fig:fig1}
    \end{figure}

    The proposed framework, illustrated in Figure \ref{fig:fig1}, includes four primary components: (1) Object Detection \&\ Tracking, (2) Data Pre-processing, (3) Model Development, and (4) Model Evaluation. Firstly, in the Object Detection \&\ Tracking phase, the YOLOv10n model is used to detect both workers and panels in the scene. Using the results from the YOLOv10n model, a DeepSORT algorithm is used to track workers and panels. Secondly, data pre-processing steps are implemented in order to modify the data to match the input format of the proposed model. Thirdly, the model development is introduced with the two proposed models. Fourthly, the model is evaluated using specific metrics. Each component will be explained in detail in the following sections.
    \unskip
    \subsection{Object detection and tracking}\label{subsec:Proposed framework}
    To automate the detection and tracking of workers and panels, the YOLOv10n model was trained on a real-world construction dataset. YOLOv10n is an object detection algorithm that was published by \cite{YOLOv10}. This model is commonly used in object detection applications as it provides a real time, efficient, and accurate object detection \cite{YOLOv10}. YOLOv10 has many model types based on its size. In this paper, we used YOLOv10n, which is a smaller model with about 2.3 million parameters. In addition to using YOLOv10n, the DeepSORT algorithm \cite{depsort} was utilized to add another parameter to the dataset, which is the tracking ID. This algorithm is normally used with other object detection models, such as YOLOv10n to add a tracking feature to the overall system. This will help to prepare the data to match the input format. 
    \subsection{Data pre-processing}\label{subsubsec:Proposed framework}
    In this step, the acquired data from step one is transformed into the following format:
                            \begin{center}
                            [Frame number, Class, ID, \(pos_x\), \(pos_y\)]
                            \end{center}
                            
    \noindent Where Frame number represent the current frame number, Class represents the class of the detected object ( i.e. Worker or Obsticale), ID represents the ID of the current tracked object, \(pos_x\) represents the x-coordinates for the tracked object, and \(pos_y\) represents the y-coordinates for the tracked object.
    The output from YOLOv10n (for object detection) and DeepSORT (for object tracking) provides object positions in terms of pixel coordinates in the image. However, for applications that require spatial reasoning or real-world measurements, it is crucial to convert these pixel coordinates to real-world coordinates.
    \par
    This conversion can be achieved by estimating the Homography Matrix H using four known real world coordinates on the site-in our case an estimated coordinates of a panel corners in the facility and the corresponding pixel coordinates are used. The function cv2.getPerspectiveTransform() in OpenCV is used to compute the homography matrix. Once the homography matrix is obtained, the pixel coordinates can be mapped to their corresponding real-world positions.
    \subsection{Model development}\label{subsubsec:Proposed framework}
    Inspired by the work of \cite{FRANCO2023109372}, the proposed model utilizes the transformer encoder-decoder architecture. The main components of transformer models include the \textit{Encoder}, \textit{Decoder}, \textit{Attention Mechanism}, \textit{Multi-head Attention}, and \textit{Positional Encoding}.

        The primary function of the encoder is to encode the source inputs into representations that can later be decoded into meaningful target sequences. The encoder consists of \(N\) identical layers (commonly \(N=6\)), each of which includes two main components: \textit{Multi-head Attention} and \textit{Fully Connected Feed-Forward Network (FFN)}. These components are connected by residual connections, which add the input of each layer to its output, helping to stabilize the training \cite{vaswani2017attention}. Layer normalization is then applied to the output of each sub-layer to ensure smooth gradient flow. 
  
        The decoder generates the target sequence by utilizing the representations produced by the encoder. Its structure is similar to that of the encoder but includes an additional attention module. The first attention module processes the encoded information from the encoder, while the second attention module applies masking to the output embeddings, ensuring that the model attends only to previous tokens in the sequence. This masking helps preserve the autoregressive property of sequence generation, where the model generates one token at a time based on the preceding context.

        In the attention modules, the attention mechanism is used to help the model focus on relevant parts of the input sequence. The attention scores are computed based on \textit{query (Q)}, \textit{key (K)}, and \textit{value (V)} matrices using the equation \ref{eq:eq11}.
            \begin{linenomath}
            \begin{equation}
                \text{Attention}(Q, K, V) = \text{softmax}\left(\frac{QK^T}{\sqrt{d_k}}\right) V
                \label{eq:eq1}
            \end{equation}
            \end{linenomath}
            
        Here, \(d_k\) represents the dimension of the key vector. The softmax function ensures that the attention scores are normalized into a probability distribution, allowing the model to weigh different parts of the input sequence appropriately.
        
        To further enhance the model's ability to focus on various aspects of the input, transformers employ \textit{multi-head attention}, where multiple attention mechanisms run in parallel. Each head attends to different subspaces of the input, allowing the model to capture various aspects of the input sequence. The outputs from all attention heads are concatenated and passed through a linear transformation to integrate the information.
        
        Since transformers do not have an inherent mechanism to process sequential order, \textit{positional encoding} is added to the input embeddings to encode the position of each token in the sequence. The positional encoding function is defined by equation \ref{eq:eq2} and \ref{eq:eq3}.
        \begin{linenomath}
        \begin{equation}
            PE_{(pos, 2i)} = \sin\left(\frac{pos}{10000^{\frac{2i}{d}}}\right)
            \label{eq:eq2}
        \end{equation}
        
        \begin{equation}
            PE_{(pos, 2i+1)} = \cos\left(\frac{pos}{10000^{\frac{2i+1}{d}}}\right)
            \label{eq:eq3}
        \end{equation}
        \end{linenomath}

        \noindent Where \(pos\) is the position of the token in the sequence, \(i\) is the dimension index, and \(d\) is the model dimension. This encoding ensures that each token's position is uniquely represented in the model's input embeddings, allowing the transformer to incorporate the sequential nature of the data. 
    
    However, to better understand the relationships between workers, GAT \cite{veličković2018graphattentionnetworks}, was implemented to model the social interactions between them. GAT is used on graph-based data, which uses self-attention to calculate the importance of each neighbor on the target node. GAT have two main components: (1) Nodes, and (2) Edges. where nodes can represent individual points in the data and the edges represent the relationship between the target node and its neighbor. \par 
    Now lets assume we have a graph \( G = (V, E) \) , where \( V \),\( E \) represents the set of nodes (i.e. workers) and set of edges (i.e. Euclidean distance), respectively. Each node \( i \in V \) is associated with a feature vector \( \tilde{h}_i \in \mathbb{R}^F \), where \( F \) is the feature dimension. The feature now are transformed into high level features, using linear transformation with a learnable parameter matrix W, where \( W \in \mathbb{R}^{F' \times F} \) and F' is the new dimension after transformation see equation \ref{eq:eq4}.
            \begin{linenomath}
            \begin{equation}
                \tilde{h}_i' = W \tilde{h}_i, \quad \tilde{h}_i' \in \mathbb{R}^{F'}
                \label{eq:eq4}
            \end{equation}
            \end{linenomath}

    The attention coefficient for the node pair \((i, j)\) at time step \( t \) is computed using equation \ref{eq:eq5}.
    \begin{linenomath}
    \begin{equation}
        \alpha^t_{ij} = \frac{\exp(\text{LeakyReLU}(\mathbf{a}^T [W \mathbf{h}^t_i \| W \mathbf{h}^t_j]))}{\sum_{k \in \mathcal{N}_i} \exp(\text{LeakyReLU}(\mathbf{a}^T [W \mathbf{h}^t_i \| W \mathbf{h}^t_k]))}
        \label{eq:eq5}
    \end{equation}
    \end{linenomath}
     
        \noindent Where \( a \in \mathbb{R}^{2F'} \) is the weight vector of a single-layer feedforward neural network, \( \| \) denotes the concatenation operation, \( \mathcal{N}_i \) represents the neighbors of node \( i \) on the graph, \(\exp\) denotes the exponential function, and \(\text{LeakyReLU}\) is the Leaky Rectified Linear Unit activation function \cite{nair2010rectified}. Finally, the output of a single GAT layer for a given node i at time t is defined by equation \ref{eq:eq6}.
        \begin{linenomath}
        \begin{equation}
            \hat{h}_i = \sigma \left( \sum_{j \in \mathcal{N}_i} \alpha_{ij} W \tilde{h}_j' \right)
            \label{eq:eq6}
        \end{equation}
        \end{linenomath}

        \noindent Where \( \sigma \) represents a non-linear activation function and \( \tilde{h}_j' \) represents the transformed feature vector of node \( j \).

        In the form of trajectory prediction, for a given person or obstacle \( i \), we define the observed trajectory sequences as:
        \begin{linenomath}
            \begin{equation}
            T_{\text{obs}} = \{ x_t^{(i)} \}_{t=-(T_{\text{obs}} - 1)}^{0}
            \label{eq:eq7}
            \end{equation}
        \end{linenomath}

    \noindent Where \( x_t^{(i)} \in \mathbb{R}^2 \) represents the Cartesian coordinates (position) at time \( t \), and \( T_{\text{obs}} \) is the number of observed time steps. Before feeding into the transformer, the 2D positions are projected into a higher-dimensional space \( D \):
        \begin{linenomath}
            \begin{equation}
            e_{\text{obs}}^{(i,t)} = x_t^{(i)} W_x
            \label{eq:eq8}
            \end{equation}
        \end{linenomath}

    \noindent Where \( W_x \in \mathbb{R}^{2 \times D} \) is a learned weight matrix, and \( e_{\text{obs}}^{(i,t)} \in \mathbb{R}^{D} \) is the transformed embedding of the observed position.
    Since Transformers do not have inherent order-awareness, a positional encoder is added to provide temporal context.

    Each input embedding \( e_{\text{obs}}^{(i,t)} \) is timestamped as:
        \begin{linenomath}
            \begin{equation}
            \xi_{\text{obs}}^{(i,t)} = PE + e_{\text{obs}}^{(i,t)}
            \label{eq:eq9}
            \end{equation}
        \end{linenomath}

    \noindent Where \( PE \) is the positional encoding vector of the same dimension \( D \). The positional encoding is defined in equation \ref{eq:eq2} and \ref{eq:eq3}. Once the the input embeddings are calculated for both workers and obstacles we get: \( \xi_{\text{worker,obs}}^{(i,t)} \) for worker input embeddings and \( \xi_{\text{obstacle,obs}}^{(i,t)} \) for obstacle input embeddings. Now for GAT model each agent \( i \) at time \( t \) has a corresponding node embedding \( \xi_{\text{worker,obs}}^{(i,t)} \). The GAT computes a new representation for each node using self-attention over its neighbors. First, each node embedding is transformed using a learnable weight matrix using equation \ref{eq:eq4}. The importance of a neighboring node \( j \) to node \( i \) is computed using equation \ref{eq:eq5}. The new node representation is computed as a weighted sum of its neighbors' embeddings using equation \ref{eq:eq6} and we get:
         \begin{linenomath}
        \begin{equation}
            \hat{h}_{worker,i} = \sigma \left( \sum_{j \in \mathcal{N}_i} \alpha_{ij} W \tilde{h}_j' \right)
            \label{eq:eq10}
        \end{equation}
        \end{linenomath}
        \par
    Now for the input to the transformer encoder we have \( \xi_{\text{worker,obs}}^{(i,t)} \) as the input for the worker encoder. This input sequence is passed through \( L \) Transformer encoder layers. Within each layer, the self-attention mechanism computes queries (\( Q \)),  (\( K \)), and (\( V \)) by multiplying the input sequence \( \xi_{\text{obs}}^{(i,t)} \) with learned weight matrices \( W_Q \), \( W_K \), and \( W_V \), respectively. The attention is computed using a scaled dot-product using equation \ref{eq:eq1}, and the results are weighted and aggregated through multi-head attention. The attention output is normalized using a LayerNorm operation. The normalized output is then passed through a feedforward network, where it undergoes a non-linear transformation with ReLU activation. After processing, the layer output is again normalized. This process is repeated through all layers of the encoder, resulting in the final encoded sequence \( \hat{h}_{\text{worker,encoder}}^{(L)} \). A similar process is applied to the obstacle encoder, resulting in \( \hat{h}_{\text{obstacle,encoder}}^{(L)} \). The final encoder output is shown in equation \ref{eq:eq11}. The addition of \( \hat{h}_{\text{worker,encoder}}^{(L)} \) and \( \hat{h}_{\text{worker},i} \) ensures that the same dimensions are used. The encoded sequence \( H_{\text{encoded}} \) is passed as the input to the decoder. The decoder uses this encoded representation to generate the predicted future trajectories. 
    
        \begin{linenomath}
    \begin{equation}
        H_{\text{encoded}} = \text{Concat} \left( \hat{h}_{\text{worker,encoder}}^{(L)} + \hat{h}_{\text{worker},i}, \hat{h}_{\text{obstacle,encoder}}^{(L)} \right)
        \label{eq:eq11}
    \end{equation}
    \end{linenomath}
    \subsubsection{TrajGATFormer model}\label{subsubsec:TrajGATFormer}
\begin{figure}[H]
                \centering
                \includegraphics[width=\textwidth]{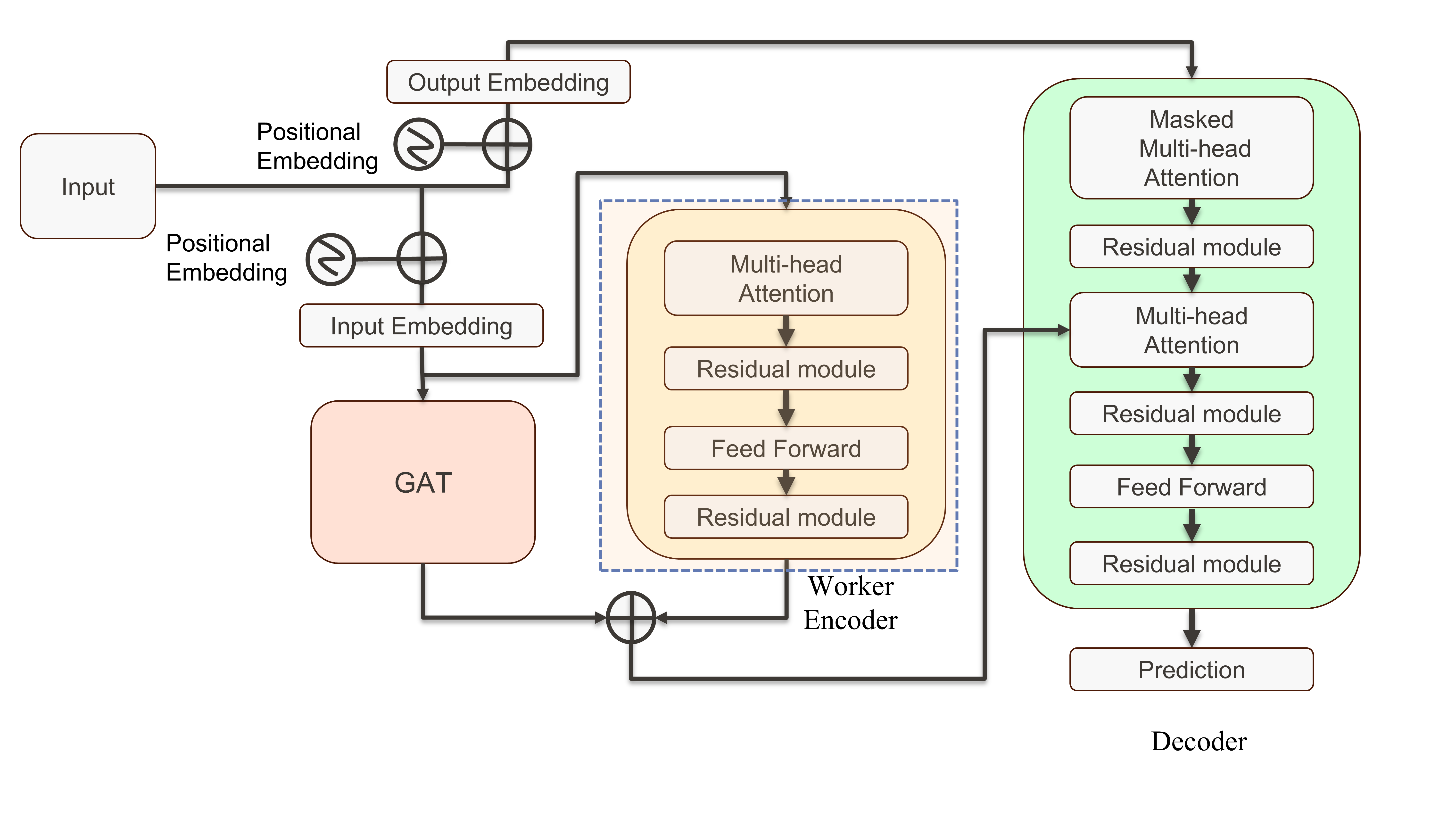}
                \caption{Overview of our TrajGATFormer worker trajectory prediction model.}
                \label{fig:model1}
            \end{figure}

    TrajGATFormer is designed to process worker trajectory information. It will take the output from equation \ref{eq:eq10} and add it to the output from the worker encoder \( \hat{h}_{\text{worker,encoder}}^{(L)} \). The results will be used as an input to the decoder. Figure \ref{fig:model1}, shows overview of TrajGATFormer model architecture.
            
    \subsubsection{TrajGATFormer-Obstacle model}\label{subsubsec:TrajGATFormer-Enhanced}
\begin{figure}[H]
                \centering
                \includegraphics[width=\textwidth]{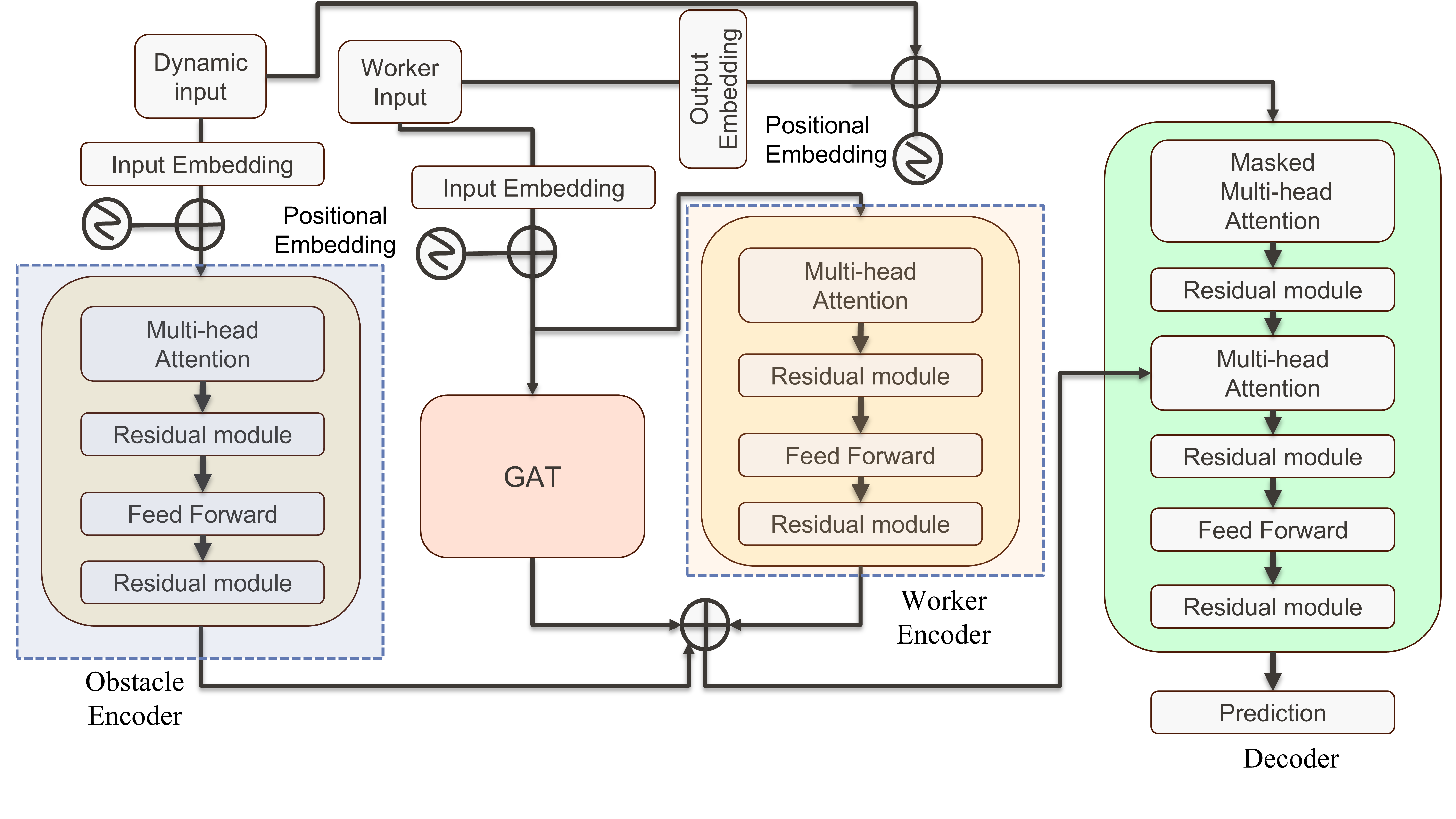}
                \caption{Overview of our TrajGATFormer-Obstacle worker and obstacle trajectory prediction model.}
                \label{fig:fig2}
            \end{figure}

    Since TrajGATFormer did not incorporate the effect of moving obstacles on the workers trajectory prediction. It only encoded the past trajectories of workers and ignored a critical object that they normally work in proximity with. Therefore, to enhance the model’s ability to represent real-world scenarios on construction sites, the proposed approach integrates information
    about moving obstacles, specifically floor or roof panels, into the trajectory prediction
    framework. The core of the model remains inspired by TrajGATFormer; however,
    a significant modification is made by introducing a dedicated encoder to process
    obstacle information. This change allows the model to account for the dynamic
    interactions between workers and obstacles, improving its accuracy and relevance in
    such environments. Figure \ref{fig:fig2}, shows overview of TrajGATFormer-Obstacle worker model architecture.

    \par
            
    \subsection{Model evaluation}\label{subsubsec:Proposed framework}
    In this section, baseline models and the evaluation metrics are discussed.
            \subsubsection{Baseline models}\label{subsubsec:Baseline models}
            In this paper different models were used to compare the performance of our model. The first model is Social-LSTM which was proposed by Alahi et al. \cite{7780479}. The second model Social-GAN which was proposed Gupta et al. \cite{gupta2018socialgansociallyacceptable}. The third and forth models are EA-Distance and EA-Direction proposed by Yang et al. \cite{buildings13061502}. The proposed model is used for worker trajectory prediction in a modular construction site, using LSTM, and it takes into account the following: (1) worker movement, (2) worker-to-worker and (3) environment-to-worker interactions. Using these models we can see whether TrajGATFormer is able to outperform them or not.
            \subsubsection{Evaluation metrics}\label{subsubsec:Baseline models}
            The evaluation metrics used in this paper are: (1) Average Displacement Error (ADE) and (2) final displacement error (FDE). ADE is defined as the mean square error between all the predicted trajectories and the corresponding ground-truths and can be calculated using equation \ref{eq:eq12}.
                \begin{linenomath}
                \begin{equation}
                        \text{ADE} = \frac{\sum_{i=1}^{N} \sum_{t=T_{\text{obs}}+1}^{T_{\text{pred}}} \left\| (\hat{x}_i^t, \hat{y}_i^t) - (x_i^t, y_i^t) \right\|}{N \times (T_{\text{pred}} - T_{\text{obs}} - 1)}
                            \label{eq:eq12}
                \end{equation}
                \end{linenomath}
                \par
                  \noindent Where \(N\) represents the number of workers or obstacle,\(T_{\text{pred}}\) denote the predicted coordinates of workers or obstacle \(i\) at the time instant t, \((\hat{x}_i^t, \hat{y}_i^t)\) denotes the predicted coordinates, \(({x}_i^t, {y}_i^t)\) denote the corresponding ground-truth coordinates, and \(\|\cdot\|\) is the Euclidean distance. FDE can be defined as the mean square error between the last coordinates of the predicted trajectories and the corresponding ground-truth and can be calculated using equation \ref{eq:eq13}. The units of both ADE and FDE are in meters as they use real-world coordinates.
                \begin{linenomath}
                \begin{equation}
                        \text{FDE} = \frac{\sum_{i=1}^{N} \left\| \left( \hat{x}_i^{(T_{\text{pred}})}, \hat{y}_i^{(T_{\text{pred}})} \right) - \left( x_i^{(T_{\text{pred}})}, y_i^{(T_{\text{pred}})} \right) \right\|}{N}
                                        \label{eq:eq13}
                \end{equation}
                \end{linenomath}

\subsection{Implementation details}\label{subsec:Implementation details}    
The implementation details of the proposed framework will be explained separately for Object detection and tracking, TrajGATFormer, and TrajGATFormer-Obstacle. However, the same hardware were used for training all of them. The hardware used for training is NVIDIA RTX 3060 GPU with 12 GB of memory, an AMD Ryzen 9 5900 12-Core Processor at 3.00 GHz, and 32 GB of RAM. The datasets used were split into three subsets, with a split of 70\%, 20\%, and 10\% for training, validation, and testing, respectively.
        \par
        First, the object detection model YOLOv10n was trained using the construction dataset with around 9288 frames for 400 epochs, batch size of 4, image size of 640 pixles, and Adam optimizer. 
        \par
         Second, the TrajGATFormer worker trajectory prediction model. A summary of the model hyper-parameters is shown in Table \ref{tab:parameters}. Regarding the \(d_{\text{model}}\), its value follows the proposed value in \cite{vaswani2017attention}, which is 512. However, for the Worker \(d_{k}\) value is 256, as proposed in several papers \cite{yuan2021agentformeragentawaretransformerssociotemporal}. 
            
            \par
            
            \begin{table}[ht]
                \centering
                \caption{\centering TrajGATFormer Model Parameters}
                \begin{tabularx}{\textwidth}{Xc}
                \toprule
                \textbf{Model Parameter} & \textbf{Value}  \\ 
                \midrule
                Worker Encoder Layers & 1  \\ 
                Decoder Layers & 1  \\ 
                \(d_{\text{model}}\) & 512  \\ 
                Attention Heads & 8 \\ 
                Worker \(d_{k}\) & 256  \\ 
                Dropout & 0.1  \\ 
                \bottomrule
                \end{tabularx}
        
                \label{tab:parameters}
            \end{table}
            
            The loss function used in the model is the L2 loss between the \((x_{\text{pred}}, y_{\text{pred}})\) coordinates predicted by the model and the real \((x_{\text{real}}, y_{\text{real}})\) coordinates. Table \ref{tab:Oparameters1} contains the optimizer parameters. The Adam optimizer was chosen for this model with 4000 warm-up steps for the learning rate; after that, it starts to decay. The learning rate equation used in the proposed model is as follows:
            \begin{equation}
            \text{Learning Rate} = \text{factor} \times \frac{1}{\sqrt{\text{\(d_{\text{model}}\)}}} \times \min\left( \frac{1}{\sqrt{\text{step number}}}, \frac{\text{step number}}{\text{warmup steps}^{1.5}} \right)
            \label{eq:learning_rate}
            \end{equation}
            Similarly to what was proposed in \cite{vaswani2017attention}, \(\beta_1\) and \(\beta_2\) are 0.9 and 0.98, respectively as values for the optimizer. The training of the models was done in 600 epochs and the batch size used is 4, as the GPU runs out of memory if the batch size is increased. A summary of TrajGATFormer and TrajGATFormer-Obstacle optimizer parameters is shown in Table \ref{tab:Oparameters1}.

             \par
            Transfer learning was investigated to see whether it improves the performance of the model or not. Two models were produced sharing the same model architecture and vary by training process. The first model was initially trained using the ETH dataset, in order to initialize the parameter weights, resulting in a base model. Then retrained using the construction dataset. The second model was trained using only the construction dataset without applying transfer learning. The first model will use the trained parameters of the base model to initialize its weights; no weight randomization is needed.

            \begin{table}[ht]
                \centering
                \caption{\centering Proposed models (TrajGATFormer and TrajGATFormer-Obstacle) Optimizer Parameters}
                \begin{tabularx}{\textwidth}{l c c}
                \toprule
                \textbf{Optimizer Parameter} & \textbf{Base model} & \textbf{\shortstack{Proposed models}} \\ \midrule
                Optimizer                    & Adam & Adam          \\ 
                Initial Learning Rate        & 0    & 0           \\ 
                Learning Rate Warm-up Steps  & 4000 & 4000         \\ 
                Scaling Factor               & 1   & 1           \\ 
                \(\beta_1\)                  & 0.9  & 0.9          \\ 
                \(\beta_2\)                  & 0.98  & 0.98         \\ 
                \(\epsilon\)                 & \(1 \times 10^{-9}\) & \(1 \times 10^{-9}\) \\ 
                Batch Size                   & 32     & 4        \\ \bottomrule
                \end{tabularx}
    
                \label{tab:Oparameters1}
            \end{table}
            \par
                        \begin{table}[!ht]
                \caption{TrajGATFormer-Obstacle Model Parameters}
                \centering
                \begin{tabularx}{\textwidth}{Xc}
                    \toprule
                    \textbf{Model Parameter} & \textbf{Value}  \\
                    \midrule
                    Worker Encoder Layers & 1  \\ 
                    Panel Encoder Layers & 1  \\ 
                    Decoder Layers & 1  \\ 
                    \(d_{\text{model}}\) & 512  \\
                    Attention Heads & 8 \\ 
                    Worker \(d_{k}\) & 256  \\ 
                    Dropout & 0.1  \\ 
                    \bottomrule
                \end{tabularx}
                \label{tab:parameters_model2}
            \end{table}
                         

             Third, the TrajGATFormer-Obstacle worker and obstacle trajectory prediction model. A summary of the model parameters used is shown in Table \ref{tab:parameters_model2}. The training of the models was done in 600 epochs, and the batch size used is 3 for workers and 1 for panels. For the observation and prediction window, it is commonly used in the literature to use an observation length of 8 frames and predict the future 12 frames.
                        \par           
\section{Datasets}\label{section:Datasets}
In this paper, two datasets namely ETH and construction datasets were used. Each dataset is split into three subsets, training, validation, and testing in an 80/10/10 split. The first dataset described in the following section will be used to train a base model to initialize the model weights. Then a new construction dataset is created to be used to fine-tune the base model using transfer learning.
      
    \subsection{ETH Dataset}\label{subsec:ETH}
        The name "ETH" was derived from ETH Zurich \cite{pellegrini2009}, this dataset contains two scenes in a birds eye view of RGB cameras that are usually used for surveillance purposes: (1) Eth; and (2) Hotel. ETH scene was taken from the top of the ETH main building and the annotation was done at 2.5 frames per second (fps), which means that one frame is annotated every 0.4 seconds. This part consists of a total of 365 different pedestrian trajectories. For Hotel, the scene was taken from the 4th floor of a hotel in Bahanhofstr in Zurich and the video was taken at a 25 fps but the annotation was done at 2.5 fps for consistency. The image resolution of both datasets is 640x640 pixles.

    \subsection{Construction Dataset}\label{subsec:Construction}
        The second dataset used in this paper is a real world construction dataset. The scenes were taken from a surveillance camera inside an off-site construction facility in Edmonton, Alberta. During training phases, the construction dataset is used, which contains two scenes from the same work station but working on different parts. In the first scene, 1907 frames were annotated with up to 3 workers and 1 moving obstacle. In the second scene, 409 frames were annotated with up to 3 workers and 1 moving obstacle. The first scene was used to train and test TrajGATFormer, and the second scene is used in addition to the first scene to train and test TrajGATFormer-Obstacle. As the first scene was focused mainly on the workers' movement in the factory, which is the focus of TrajGATFormer, and the second scene we tried to focus more on video portions where the obstacle is actually moving to add more data for the model to learn.

        \subsection{Statistical analysis}\label{subsec:Statistical analysis}
        To better understand the proposed construction dataset, a statistical analysis of the recorded trajectories is presented. This analysis is performed on scene 1 from the construction dataset. To analyze the dataset two factors were considered: (1) Workers mean walking speed, and (2) workers stop fraction. The mean walking speed of a trajectory is calculated by averaging the speed between consecutive points. The speed between two consecutive points is given by the Euclidean distance between those points, multiplied by fps. For a trajectory with \( n \) points, the mean speed can be calculated as follows:
        
                \[
                v_{\text{mean}} = \frac{1}{n-1} \sum_{i=1}^{n-1} \| \mathbf{p}_{i+1} - \mathbf{p}_i \| \cdot {\text{fps}}
                \]
                \par
                \noindent Where \( v_{\text{mean}} \) is the mean speed, \( \mathbf{p}_i \) is the position of the \(i\)-th point in the trajectory, \( f_{\text{rate}} \) is the frame rate (frames per second), \( \| \mathbf{p}_{i+1} - \mathbf{p}_i \| \) is the Euclidean distance between consecutive points. The stop fraction is the proportion of the total time during which the speed is considered low and the worker is deemed as stopped. To categorize whether the person is stopped researches considered using thresholds as seen in \cite{walkingspeed}, where the author used a threshold of 0.5 m/s for walking pedestrians. However, this might not be the case for walking workers in an offsite construction facility as they normally work in a set of working stations, which might reduce the overall walking speed and consonantly the stopping threshold. Therefore, we decided to use a bit lower threshold than pedestrians and used a threshold of 0.4 m/s.  The stop fraction is calculated as:
                \par
                \[
                \text{Stop Fraction} = \frac{\sum_{i=1}^{n-1} \mathbb{1}( \| \mathbf{p}_{i+1} - \mathbf{p}_i \| \cdot {\text{fps}} < {\text{thresh}} )}{n-1}
                \]
                \par
                \noindent Wher:\par
                - \( \mathbb{1}(\cdot) \) is the indicator function that is 1 if the condition inside is true (speed is below the stop threshold), and 0 otherwise.\par
                - \( {\text{thresh}} \) is the threshold speed below which the worker is considered to have stopped.\par
                \par
       \begin{figure}[H] 
                \centering
                \includegraphics[width=\textwidth]{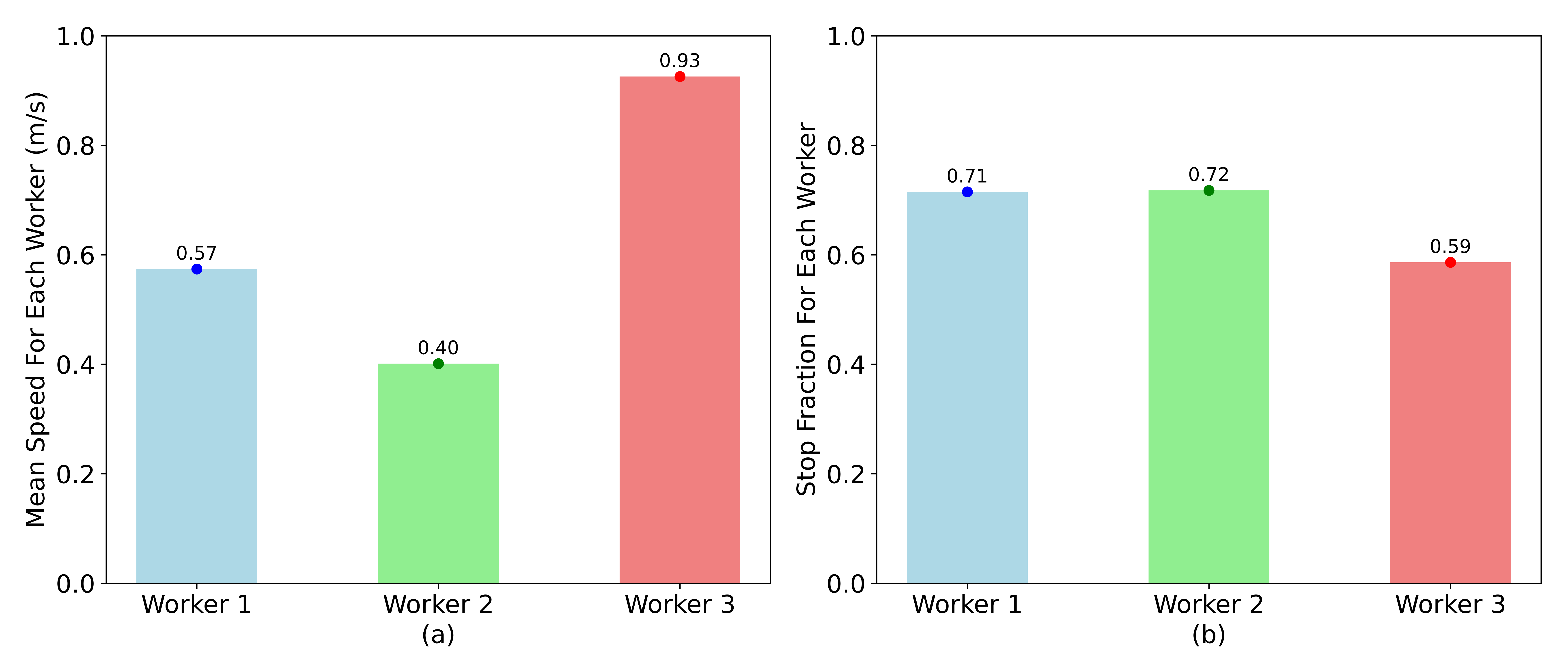}
                \caption{Statistical analysis of the construction dataset.} 
                \label{fig:dataset_analysiss}
                \par

    \end{figure}
            As we can see from Figure \ref{fig:dataset_analysiss}, the workers mean walking speed is plotted on (a) and the workers stop fraction is plotted on (b). From this figure assuming a threshold of 0.4 m/s worker 1 and 2 are mostly not moving during the recorded period. While worker 3 has been actively walking with a mean walking speed of 0.93 m/s. This is shown in the stop fraction figure, where workers 1 and 2 are considered stopped for more than 70\% of the time, while worker 3 is considered stopped for around 59\% of the time. Despite these results, it actively mirrors the real-world scenario where the workers are spending most of their time working in the working station.

\section{Results and Discussion}\label{Results}
    The results of the proposed TrajGATFormer and TrajGATFormer-Obstacle worker and/or obstacle trajectory prediction models are presented in this section.

    \subsection{Object detection and tracking results}
    The detection model was able to achieve high detection results. The results are 99.39\% for precision, 99.4\% for recall, 93.7\% for mean Average Precision (mAP), and 99.4\% for mean Average Precision at 50 (mAP50). For training, their are three training losses that was monitored: (1) Box loss, (2) Class loss, and (3) DFL loss. Despite the high performance of the detection model, the model faces challenges when exposed to drastically different environment. There are some reasons that can justify this, as when applied to different working environment several factors change, such as lighting, recording equipment, camera location.

    \subsection{Training Results of TrajGATFormer and TrajGATFormer-Obstacle}
\begin{figure}[!h] 
            \centering
                \includegraphics[width=0.8\textwidth]{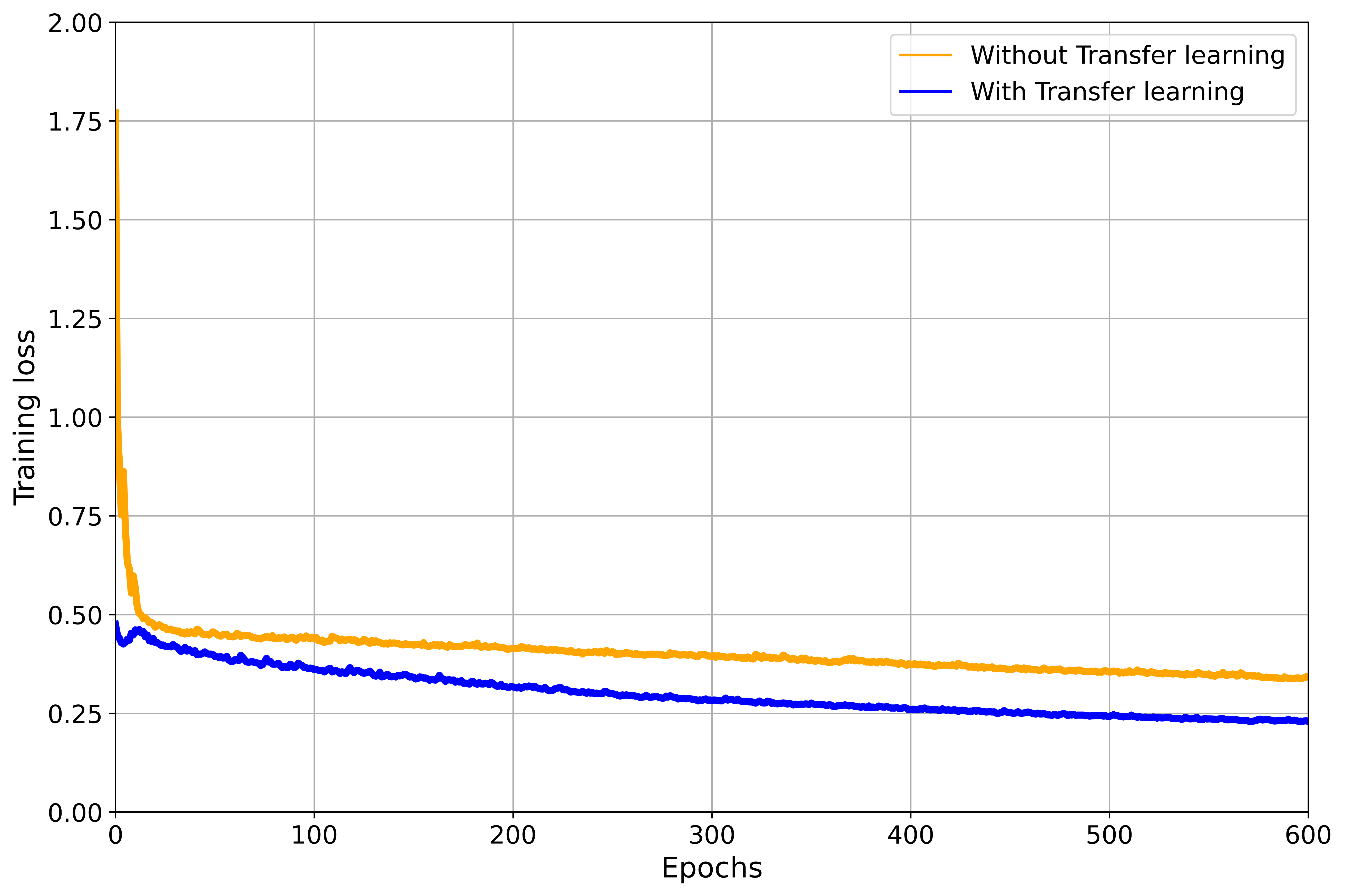}
                \caption{Training loss curves comparison between training with transfer learning and without for TrajGATFormer.} 
                \label{fig:TrajGATFormer_training_loss}
        \end{figure}

        The training results of TrajGATFormer with transfer learning show that the model in the early epoch exhibits an over-fitting signs; however, after letting the model train for more epochs the model performance improved. The comparison between TrajGATFormer with transfer learning and without transfer learning is shown in Figure \ref{fig:TrajGATFormer_training_loss}. It can be seen from the figure that training with transfer learning starts with a lower training loss compared to TrajGATFormer trained without transfer learning. However, TrajGATFormer trained without transfer learning was able to converge very fast within a few epochs and followed the training loss of the other model. From this figure, we can see that the model with transfer learning resulted in overall lower training error.
\par
            Transfer learning was not investigated TrajGATFormer-Obstacle, as using transfer learning requires the use of the same tensor sizes before and after and a similar model architecture. However, in this case, a new parameter is added, which is the panels, and a new encoder is added too.
    

    \subsection{Performance Evaluation of TrajGATFormer and TrajGATFormer-Obstacle}
        The performance of the proposed models will be measured and compared to the baseline models introduced earlier. We kept the same test setting, evaluation metrics, observation and prediction windows, batch size, and number of epochs during training. The observation window will be 8 frames and the model predicts the next 12 frames. The batch size and the epochs used are 4 and 600, respectively.
        \par
        Table \ref{tab:worker_results_model2}, shows a summary of the worker trajectory results from all models. While table \ref{tab:panel_results_model2}, shows the trajectory predictions of the moving obstacle in our case a moving panel. The presented results showed that TrajGATFormer outperformed all baseline models with an improvement percentage range of approximately [27\% - 31\%] for ADE and a range of approximately [26\% - 29\%] for FDE. However, TrajGATFormer-Obstacle results showed a significant improvement compared to baseline models, up to more than 40\% improvement in ADE and 38\% in FDE  compared to LSTM. In addition, when comparing TrajGATFormer-Obstacle and TrajGATFormer, the results showed an improvement. TrajGATFormer-Obstacle has an ADE of 1.03, while TrajGATFormer has an ADE of 1.25. This represents an improvement of about 17.6\% in TrajGATFormer-Obstacle, which means it is more accurate in predicting the overall trajectory. TrajGATFormer-Obstacle has an FDE of 2.06, compared to 2.35 for TrajGATFormer. This results in an improvement of around 12.34\% in terms of the accuracy of the final prediction.
    
        \begin{table}[h]
            \caption{Results For Worker Trajectories}
            \begin{tabularx}{\textwidth}{bss}
            \toprule
                \textbf{Model Name} & \textbf{ADE} & \textbf{FDE} \\ 
                
                \midrule
                LSTM & 1.753 & 3.332 \\ 
           
                SGAN & 1.767 & 3.343 \\
          
                EA-Distance & 1.825 & 3.349 \\
        
                EA-Direction & 1.734 & 3.201 \\ 
                TrajGATFormer & 1.250 & 2.350 \\
        
                TrajGATFormer-Obstacle & 1.030 & 2.060 \\ 
        \bottomrule
            \end{tabularx}
            \label{tab:worker_results_model2}
        \end{table}
        
\begin{figure}[h]
                    \centering
                    \includegraphics[width=\textwidth, height=0.4\textheight, keepaspectratio]{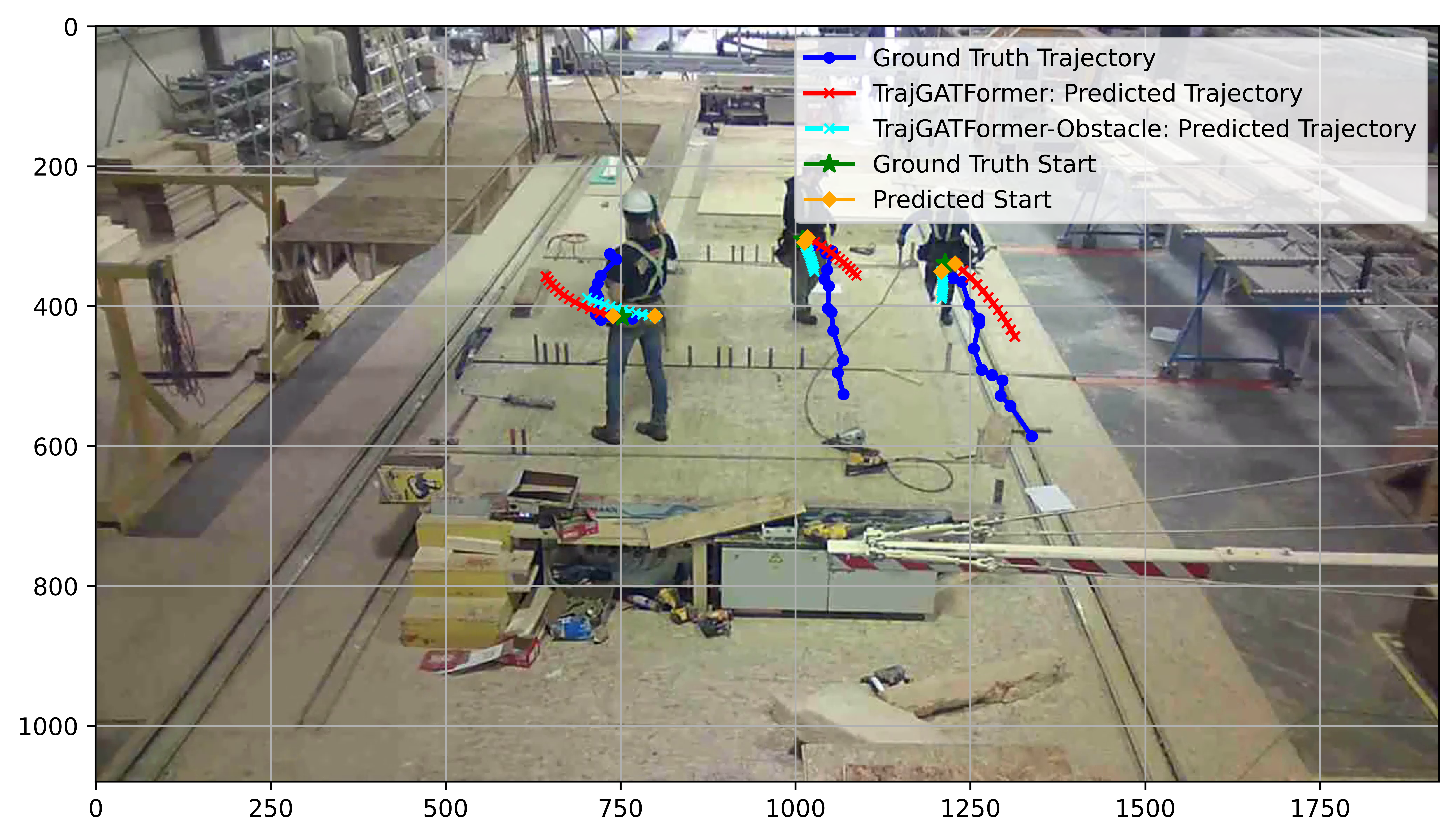} 
                    \vspace{0.5cm} 
                    \hspace{0.5cm}
                    \includegraphics[width=\textwidth, height=0.4\textheight, keepaspectratio]{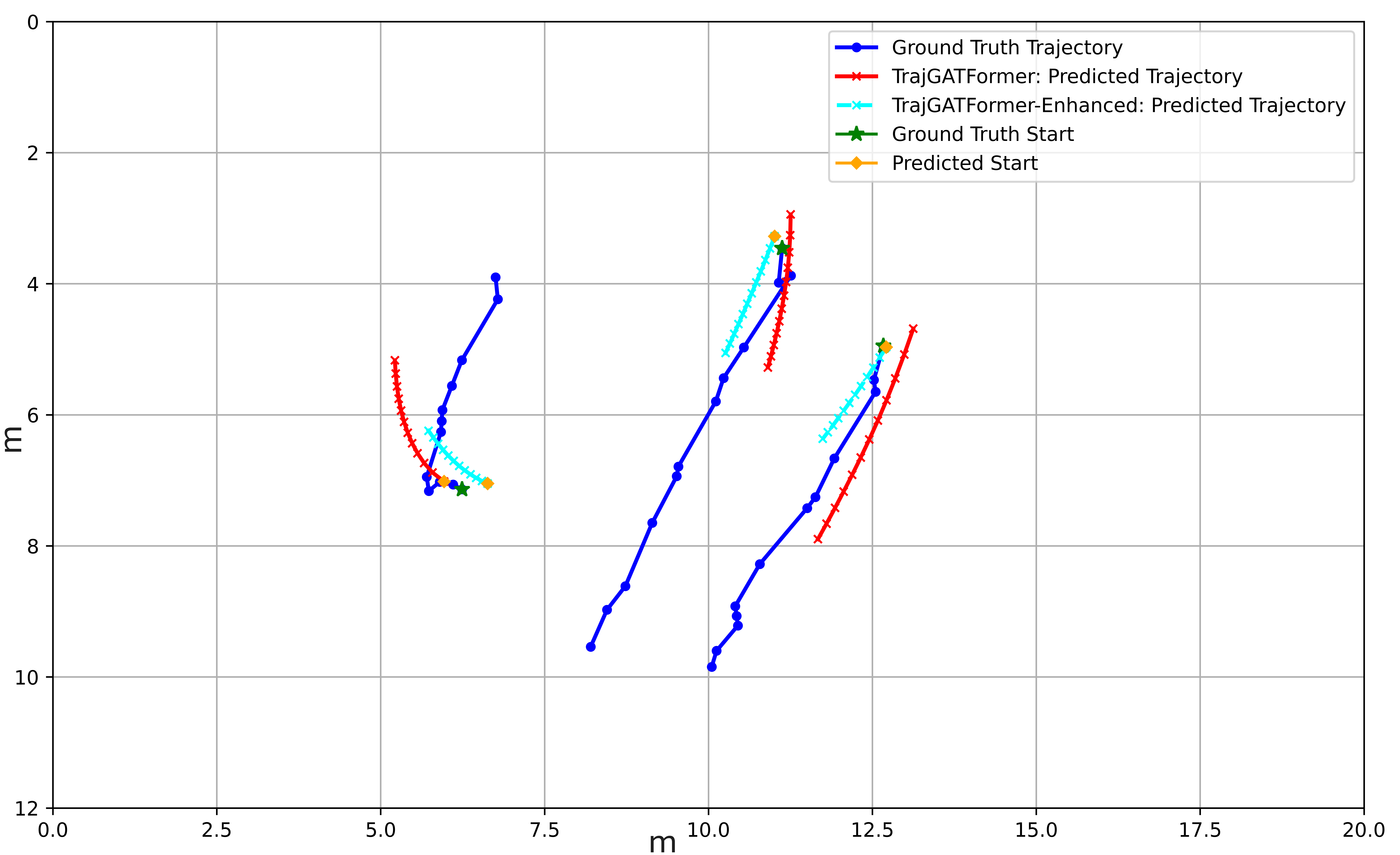} 
                    \caption{A comparison between TrajGATFormer and TrajGATFormer-Obstacle predicted vs ground truth trajectories.}
                    \label{fig:TrajGATFormer_comp}
                \end{figure}

\begin{figure}[h!]
              \centering
              \newcommand{\subwd}{0.45\textwidth}
              \newcommand{\subht}{5cm}
            
              \begin{subfigure}[b][\subht][c]{\subwd}
                \centering
                \includegraphics[width=\textwidth,height=\subht,keepaspectratio]{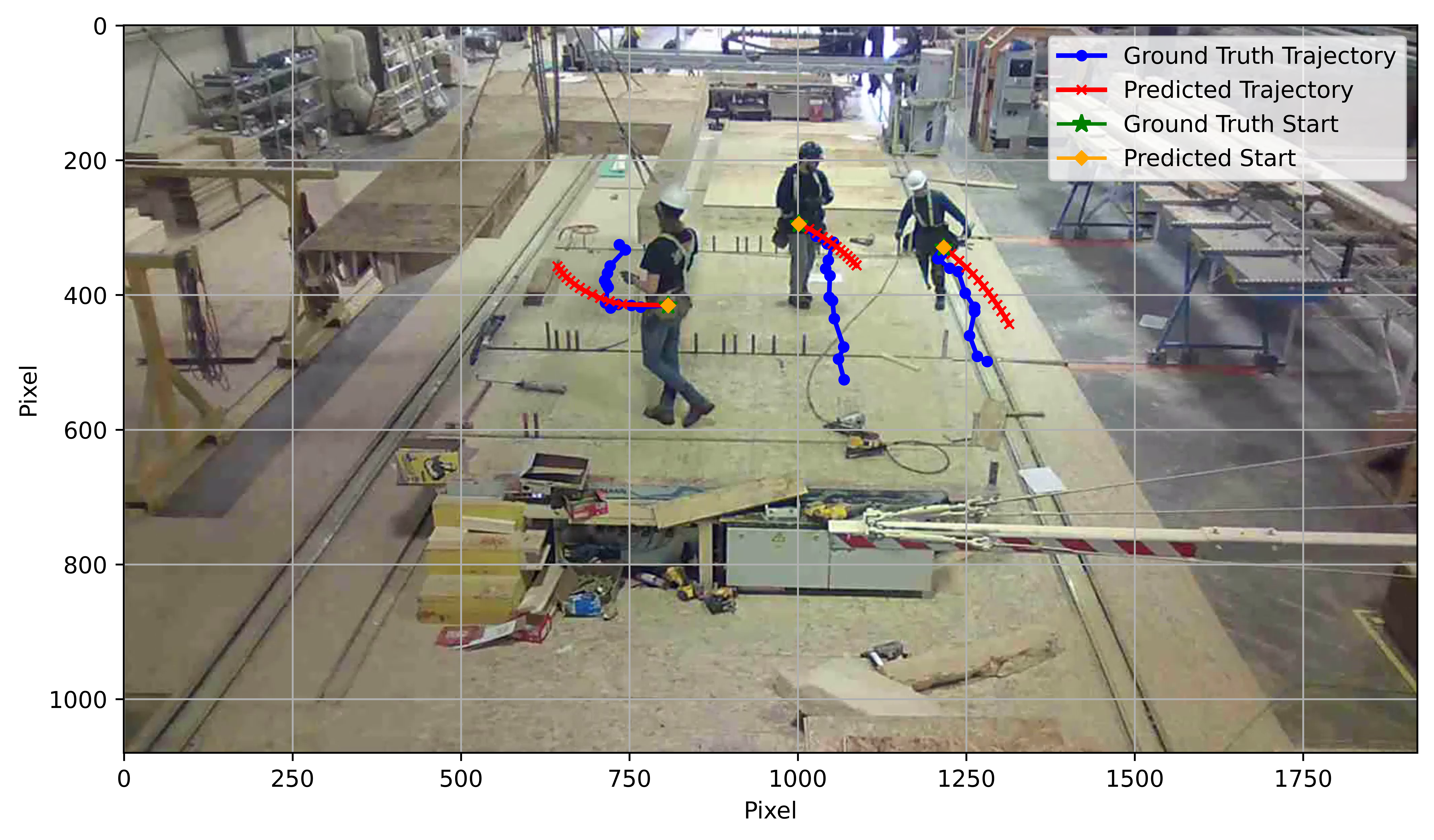}
                \caption{}
                \label{fig:trajgatformer-img}
              \end{subfigure}
              \hfill
              \begin{subfigure}[b][\subht][c]{\subwd}
                \centering
                \includegraphics[width=\textwidth,height=\subht,keepaspectratio]{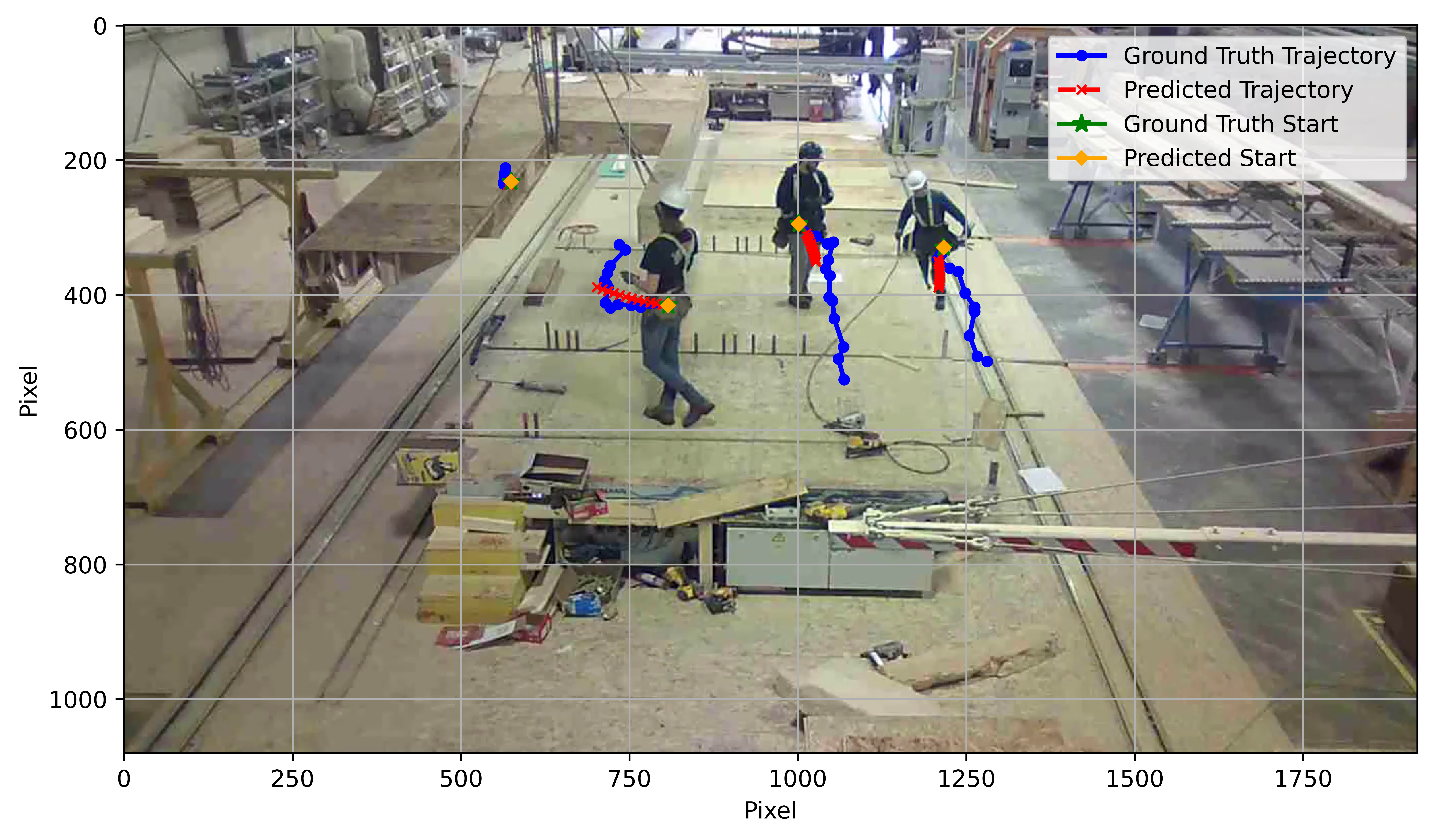}
                \caption{}
                \label{fig:trajgatformerobs-img}
              \end{subfigure}
            
              \vspace{0.2cm}
            
              \begin{subfigure}[b][\subht][c]{\subwd}
                \centering
                \includegraphics[width=\textwidth,height=\subht,keepaspectratio]{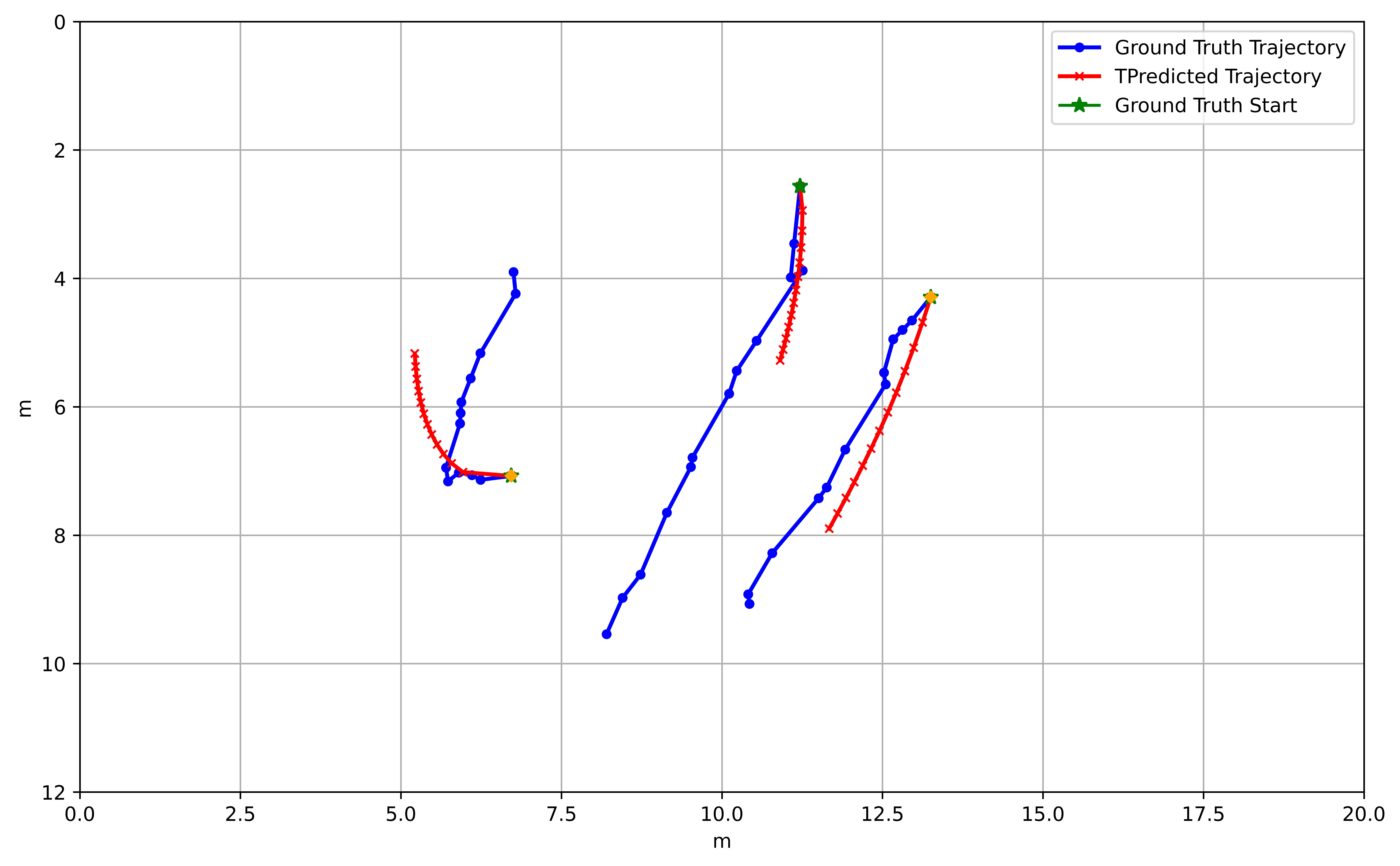}
                \caption{}
                \label{fig:trajgatformer-nobg}
              \end{subfigure}
              \hfill
              \begin{subfigure}[b][\subht][c]{\subwd}
                \centering
                \includegraphics[width=\textwidth,height=0.2\textheight]{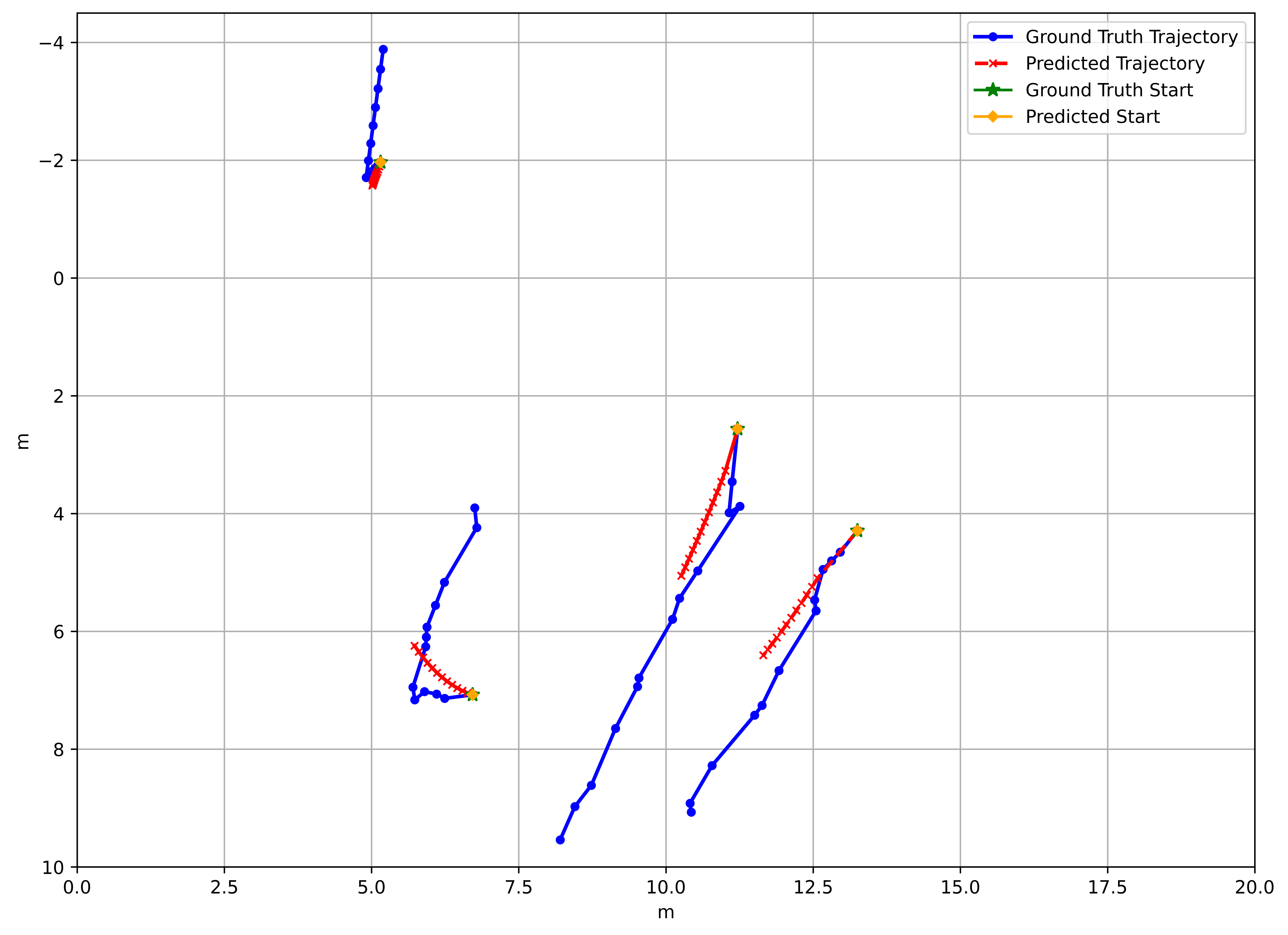}
                \caption{}
                \label{fig:trajgatformerobs-nobg}
              \end{subfigure}
            
              \caption{Predicted vs.\ ground-truth long trajectories: \subref{fig:trajgatformer-img}–\subref{fig:trajgatformerobs-nobg}. The yellow and green markers denote the starting point of the predicted and ground-truth trajectories, respectively.}
              \label{fig:TrajGATFormer_long}
            \end{figure}

        \begin{table}[h]           
                \caption{Results For Panel Trajectories}
                \begin{tabularx}{\textwidth}{XXX}
                \toprule
                \textbf{Model Name} & \textbf{ADE} & \textbf{FDE} \\
                \midrule
                TrajGATFormer-Obstacle & 1.630 & 3.790 \\ 
                \bottomrule
                \end{tabularx}
             \label{tab:panel_results_model2}
        \end{table}      
        Figure \ref{fig:TrajGATFormer_comp}, compares the predicted trajectories between TrajGATFormer and TrajGATFormer-Obstacle. The predicted trajectories are shown for the same set of frames (i.e., they share the same frame for each trajectory prediction point). The results showed improvement for all workers. However, it showed that the TrajGATFormer-Obstacle results follow the general direction of the ground-truth but have shorter trajectories. This can be seen in the trajectories of the worker on the left side. Figure \ref{fig:TrajGATFormer_long}(a), shows a comparison between a prediction made by TrajGATFormer for a time stamp and the corresponding ground-truth trajectories for long trajectories. While \ref{fig:TrajGATFormer_long}(c) shows the trajectories in the form of real-world coordinates. The results demonstrate that the model effectively captures the general movement patterns of workers, even in instances of irregular motion, such as the worker on the left side. However, in some cases, the predicted trajectory lengths deviate from the ground-truth. This discrepancy may stem from the dataset's inherent structure. Unlike pedestrians, workers exhibit distinct movement patterns, often pausing for extended periods to complete tasks before proceeding to their next location. 
        \clearpage

            \begin{figure}[h!]
                          \centering
                          \begin{subfigure}[b]{0.45\textwidth}
                            \includegraphics[width=\textwidth, keepaspectratio]{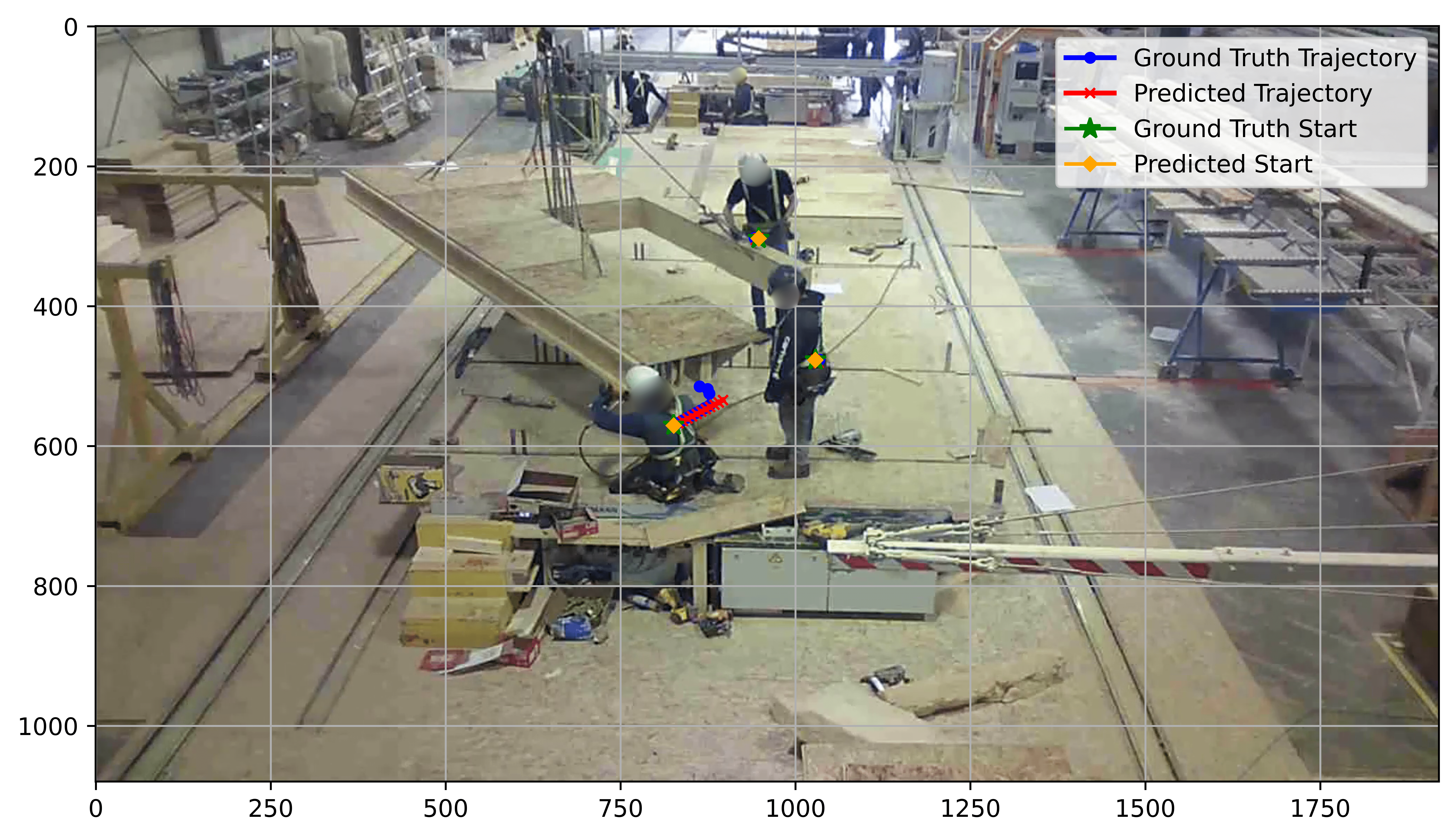}
                            \caption{ }
                          \end{subfigure}
                          \hfill
                          \begin{subfigure}[b]{0.45\textwidth}
                            \includegraphics[width=\textwidth, keepaspectratio]{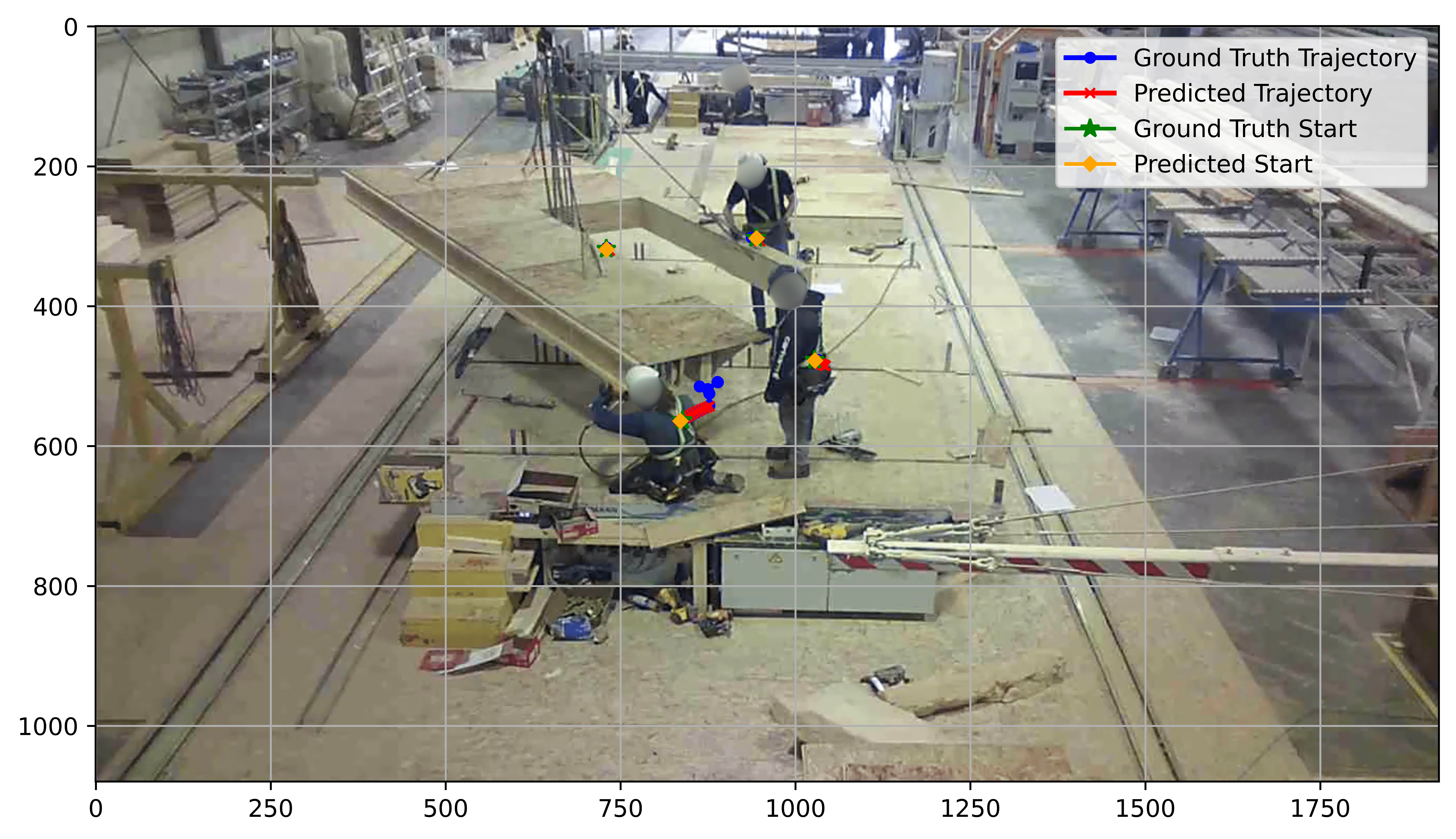}
                            \caption{ }
                            
                          \end{subfigure}
                        \vspace{0.1cm}
                          \begin{subfigure}[b]{0.45\textwidth}
                            \includegraphics[width=\textwidth, keepaspectratio]{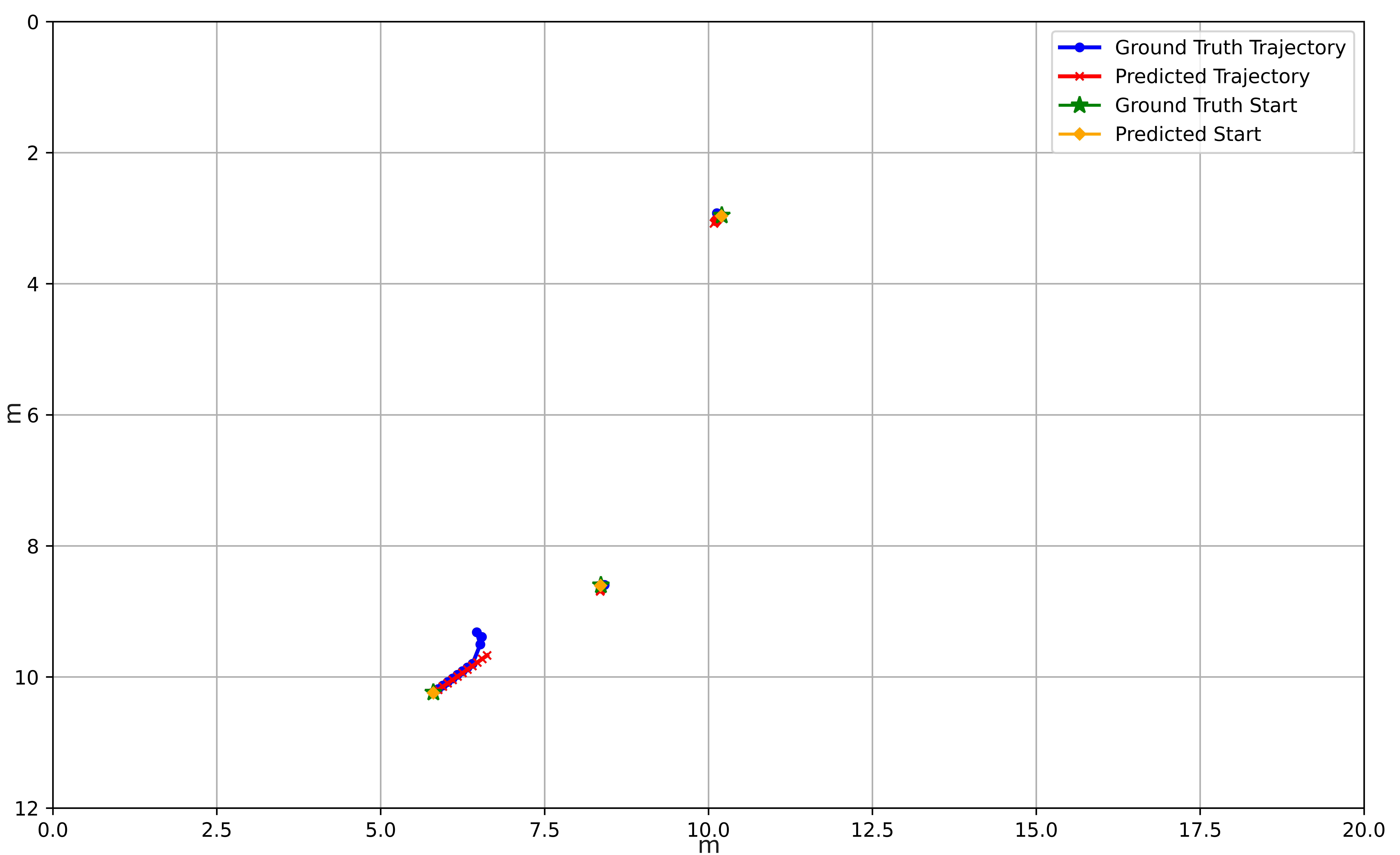}
                            \caption{ }
                            \label{fig:long-bg}
                          \end{subfigure}
                          \hfill
                          \hspace{0.2cm}
                          \begin{subfigure}[b]{0.45\textwidth}
                            \includegraphics[width=\textwidth, keepaspectratio]{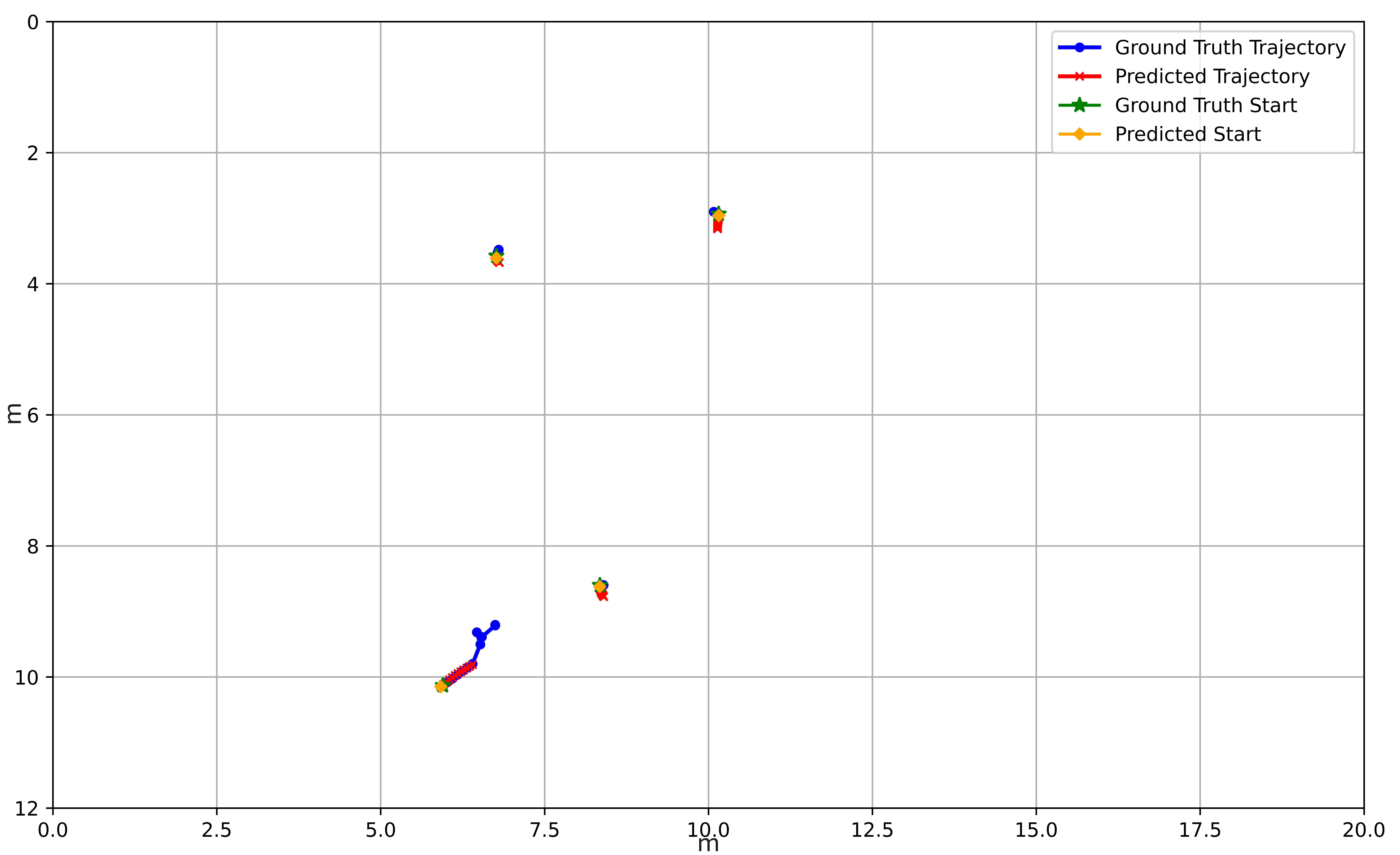}
                            \caption{ }
                          \end{subfigure}
                        
                          \caption{Predicted vs.\ ground truth short trajectories for: (a) TrajGATFormer, and (b) TrajGATFormer-Obstacles.}
                          \label{fig:TrajGATFormer_all}
                    \end{figure}

       Furthermore, Figure \ref{fig:TrajGATFormer_long}.b illustrates the actual trajectories of three workers and one moving panel, along with their predicted trajectories and their corresponding real world coordinates are shown in \ref{fig:TrajGATFormer_long}(d). The overall performance of TrajGATFormer-Obstacle is better compared to TrajGATFormer. However, the model still suffers from long trajectories, as seen in Figure \ref{fig:TrajGATFormer_long}(b) and (d). It's clear that the model can predict the direction of the movement accurately, but it can not predict the full length of the ground truth, as shown in the trajectories of workers in the middle and right sides. In addition, the model still suffers from the sudden or sharp turns of the workers, as shown in the trajectory of the worker on the left side. For the panel trajectory, the model can slightly follow the length and direction of long step movements. However, the trajectories are much worse than those of the workers, as shown in Table \ref{tab:panel_results_model2}. For panels, the poor performance can be explained by the limited annotated data that includes moving objects. However, when checking the performance of the models for short step movements, the models were able to accurately predict the direction and length of the trajectories as shown in Figure \ref{fig:TrajGATFormer_all}. In addition, the trajectory of the worker on the left side shows that the ground truth trajectory has a sudden turn, which caused the poor performance of the predicted trajectory, even though in the earlier steps, the predicted trajectories follow the ground truth until the sudden turn. 
        \par

\section{Conclusions}\label{Conclusions}
In conclusion, to mitigate hazards in offsite construction where workers operate in crowded environments with both static and dynamic obstacles, accurate trajectory prediction is essential for developing effective struck-by alarm systems. Traditional approaches have largely overlooked the dynamic nature of obstacles, limiting their ability to capture workers movements. In response, this paper proposes two trajectory prediction models, TrajGATFormer and TrajGATFormer-Obstacle. The first part of the work focused on the development of TrajGATFormer, which demonstrated improved prediction of worker trajectories using transformer-based layers and GATs. This model effectively captured the temporal and spatial dependencies of worker movements, significantly reducing prediction errors compared to traditional models. TrajGATFormer achieved a reduction in ADE by 27.92\% to 31.51\% and FDE by 26.57\% to 29.84\% compared to models like LSTM and SGAN. The second part extended the initial framework to include both workers and obstacles with TrajGATFormer-Obstacle. By incorporating an additional transformer encoder for obstacle tracking, this model further improved accuracy in predicting interactions between workers and dynamic objects. TrajGATFormer-Obstacle achieved a reduction in ADE by 17.6\% and FDE by 12.34\% compared to TrajGATFormer. While the model demonstrates significant improvements in handling short-term trajectories, performance declines for long trajectories and abrupt movement changes, highlighting the need for more diverse training data and adaptive learning strategies. Future work should focus on addressing these challenges to further refine prediction accuracy and robustness in complex construction scenarios.

\section{Limitations and Future Direction}\label{limitations}
This research, while successful in developing an advanced framework for trajectory prediction, faced several limitations that provide avenues for future improvement. First, the dataset used for model training was limited in both scope and diversity, particularly with regard to obstacles types, worker interactions, and the working environment. This constraint may impact the model's performance when deployed in highly varied or complex scenarios, where prediction accuracy for rare or complex interactions is essential. Second, while the models performed exceptionally well in short-term trajectory prediction, their predictive accuracy decreases in long-term scenarios or when workers exhibit sudden, unpredictable changes in movement or prolonged inactivity.\par
To build on the advancements made in this research, several promising directions for
future work are identified. First, expanding data collection efforts to cover a wider range of construction sites, obstacle types, and interaction scenarios would significantly improve the adaptability of the model in different environments, especially
when dealing with unusual but high-risk interactions. Second, implementing the
framework on live construction sites would provide valuable practical insights and
allow real-world testing of the performance, robustness, and responsiveness of the
model, particularly when faced with real-time constraints and hardware limitations.
Another important future direction is the integration of the prediction framework
with automated safety protocols. By linking real-time trajectory predictions to alert
or intervention systems, the framework could enable immediate responses to emerging
risks, greatly improving on-site safety.

\bibliographystyle{elsarticle-num-names} 
\bibliography{cas-refs}

\end{document}